%% file: main.tex
\documentclass[journal]{IEEEtran}
\ifCLASSINFOpdf
\else
\fi
\usepackage{graphicx} 
\usepackage{amsmath}
\usepackage{amssymb}
\usepackage{url}
\usepackage{xcolor}
\usepackage{soul}

\usepackage{comment}
\usepackage{amsmath,amssymb} 
\usepackage{color}
\usepackage{cite}
\usepackage{threeparttable}
\usepackage{comment}
\usepackage{adjustbox}
\usepackage{array,multirow}
\usepackage{colortbl}
\usepackage[percent]{overpic}

\usepackage{booktabs}
\usepackage{arydshln}

\def\eg{\textit{e.g.}}

\newcommand{\ve}[1]{\mathbf{#1}} %

\begin{document}
%
\title{Global Aggregation then Local Distribution for Scene Parsing}
%
%
%
\author{
Xiangtai Li$^*$,
Li Zhang$^*$,
Guangliang Cheng,
Kuiyuan Yang,
Yunhai Tong,
Xiatian Zhu,
Tao Xiang
\thanks{Xiangtai Li and Li Zhang contributed equally.
Xiangtai Li, Yunhai Tong are with the Key Laboratory of Machine Perception, School of EECS, Peking University.
Li Zhang is with School of Data Science, Fudan University.
Xiatian Zhu and Tao Xiang are with University of Surrey.
Kuiyuan Yang is with DeepMotion.
Guangliang Cheng is with SenseTime.
E-mail: \{lxtpku, yhtong\}@pku.edu.cn,
lizhangfd@fudan.edu.cn,
kuiyuanyang@deepmotion.ai,
guangliangcheng2014@gmail.com}
}

%
%

\markboth{Journal of \LaTeX\ Class Files,~Vol.~14, No.~8, August~2015}%
{Shell \MakeLowercase{\textit{et al.}}: Bare Demo of IEEEtran.cls for IEEE Journals}
%



\maketitle

\input{0abstract.tex}


%
\IEEEpeerreviewmaketitle

\input{1introduction.tex}
\input{2relatedwork.tex}
\input{3method.tex}
\input{4experiment.tex}
\input{5conclusion.tex}


%


\section*{Acknowledgment}
We gratefully acknowledge the support of Sensetime Research for providing the computing resources in carrying out this research. This work was supported by the National Key Research and Development Program of China (No.2020YFB2103402).

\ifCLASSOPTIONcaptionsoff
  \newpage
\fi



{
\bibliographystyle{IEEEtran}
\bibliography{IEEEabrv,egbib}
}

\end{document}

%% file: 0abstract.tex
\begin{abstract}
Modelling long-range contextual relationships is critical for pixel-wise prediction tasks such as semantic segmentation.
However, convolutional neural networks (CNNs) are inherently limited to model such dependencies due to the naive structure in its building modules (\eg, local convolution kernel).
While recent global aggregation methods are beneficial for long-range structure information modelling, they would oversmooth and bring noise to the regions contain fine details (\eg,~boundaries and small objects), which are very much cared in the semantic segmentation task. To alleviate this problem, we propose to explore the local context for making the aggregated long-range relationship being distributed more accurately in local regions. In particular, we design a novel local distribution module which models the affinity map between global and local relationship for each pixel adaptively. Integrating existing global aggregation modules, we show that our approach can be modularized as an end-to-end trainable block and easily plugged into existing semantic segmentation networks, giving rise to the \emph{GALD} networks. Despite its simplicity and versatility, our approach allows us to build new state of the art on major semantic segmentation benchmarks including Cityscapes, ADE20K, Pascal Context, Camvid and COCO-stuff. Code and trained models are released at \url{https://github.com/lxtGH/GALD-DGCNet} to foster further research.
\end{abstract}

%% file: 1introduction.tex
\section{Introduction}

\begin{figure}
	\centering
	\includegraphics[width=0.48\textwidth]{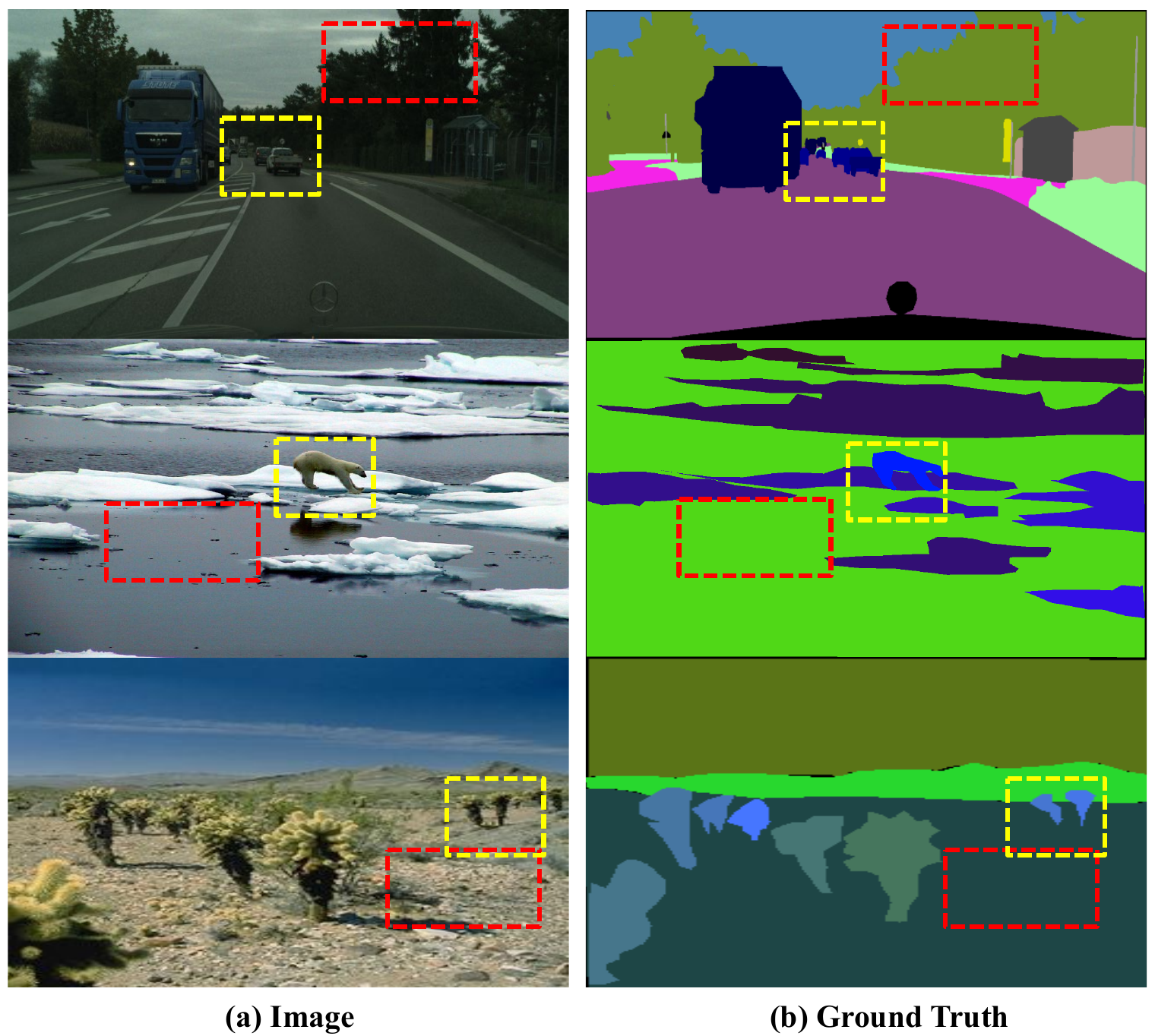}
	\caption{
		The goal of semantic segmentation is to recognize each pixel with a semantic label (\eg, sky or car).
		Despite the best efforts from the computer vision researchers, semantic segmentation remains a largely unsolved problem. This is due to the dramatic changes in object scales, occlusion and illumination make it challenging to parse large stuffs (red boxes) and small objects (yellow boxes). (a) input images, (b) ground truth.
		Best viewed in color.}

	\label{fig:teaser_1}
\end{figure}

\IEEEPARstart{S}{cene} parsing or called semantic segmentation aims at assigning a class label to each pixel in an image.
It has many applications such as auto-driving, robotic system, medical imaging and human body parsing. 
Since it requires to classify each pixel in the image, it implicitly involves pixel-level object recognition, pixel-level object localization and boundary detection.
Due to various scene and objects scales, a robust and accurate segmentation system needs to perform equally well on all of these implied tasks. 
As shown in the Figure~\ref{fig:teaser_1}, it is very challenging to parse each pixel including both large stuffs and small objects. Pixels that belong to the large object shown in red boxes or small object shown in yellow boxes all matter for final fine-grained masks. 


Recently, 
Convolutional Neural Networks (CNN) have largely improved the performance of computer vision tasks including image classification, object detection and other low level vision tasks like image deblurring. 
Applying CNN to dense prediction tasks like semantic segmentation through fully convolutional networks (FCNs) has made big progress. 
Nonetheless, there are still several problems and limitations which prevent obtaining higher quality segmentation masks. 
One of the most important aspects is about 
{\em context modeling}. The recent state-of-the-art approaches mainly focus on how to learn better global context information since stacking more convolutional layers is not an effective strategy to achieve large receptive fields for long-range dependency modeling~\cite{zhou2014object,luo2016understanding}. 
We call these approaches as {\em Global Aggregation} (GA) operators.  
In contrast to the standard convolutional layer which aggregates features in a small local window, GA modules use long-range operators such as averaging pooling~\cite{pspnet, deeplabv3} and spatial-wise feature propagation over the whole image space~\cite{Nonlocal,ocnet,DAnet,emanet}. 
GA modules have consistently improved FCNs especially for large objects in the scene.

\begin{figure}
	\centering
	\includegraphics[width=0.45\textwidth]{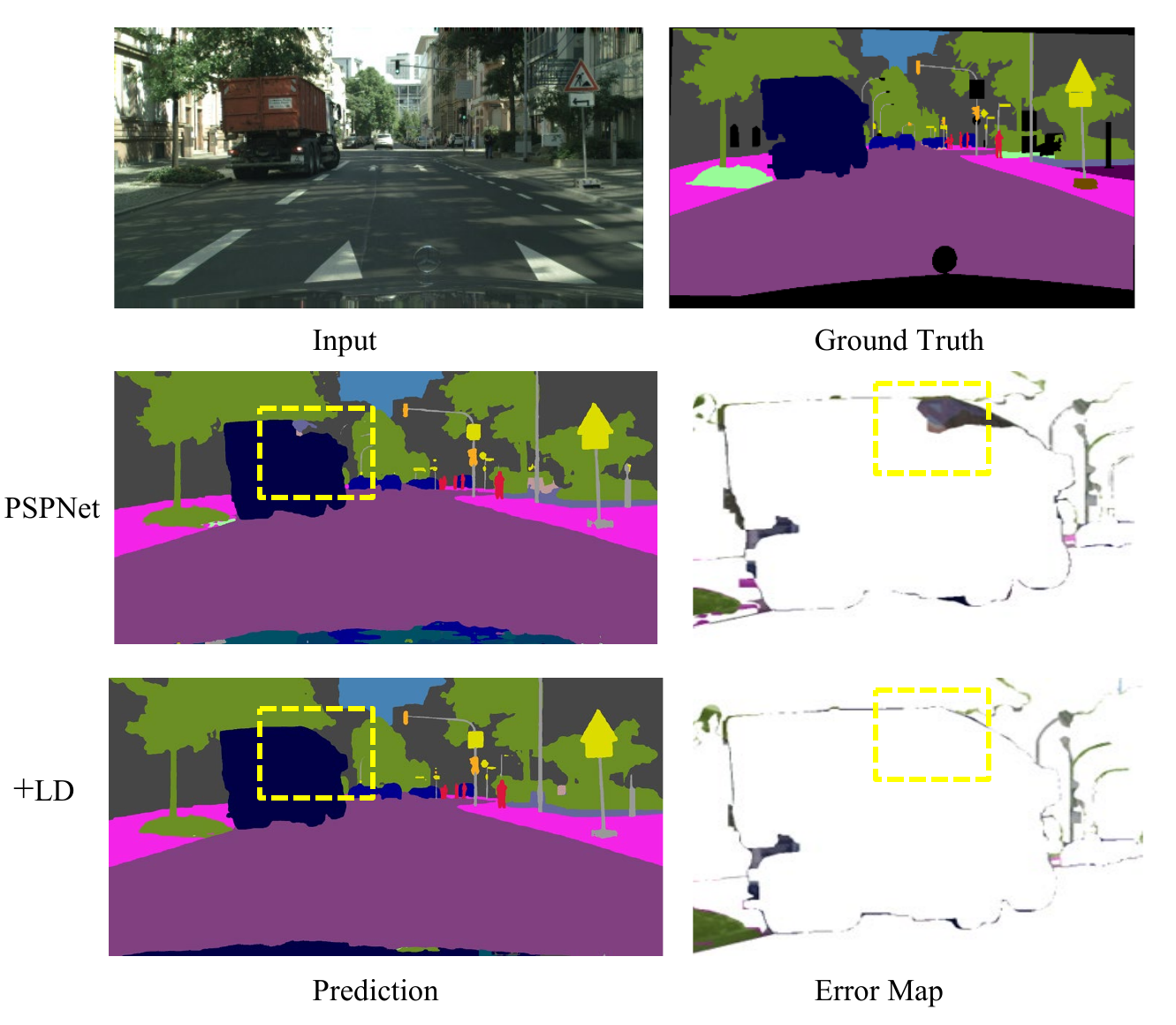}
	\caption{
	Visualizing the effect of our Local Distribution (LD) module on top of 
	state-of-the-art PSPNet \cite{pspnet}
	in reducing the error of segmenting
	object boundaries.
	Best viewed in color.
	}
	\label{fig:teaser_2}
\end{figure}

Meanwhile, 
GA modules are inferior for reasoning small patterns such as object boundaries and small objects. This is because the features yielded by GA modules are often over-smooth and
therefore hurt the inference on small objects.
More specifically, over-emphasizing global context leads to inferior results on small objects where the local context information is overwhelmed by dominant global context. For example, as shown in Figure~\ref{fig:teaser_2}, 
the truck's top boundary was segmented
mistakenly by the state-of-the-art model PSPnet~\cite{pspnet}
due to applying a coarse pyramid pooling module.

\begin{figure*}
\centering
\includegraphics[width=\textwidth]{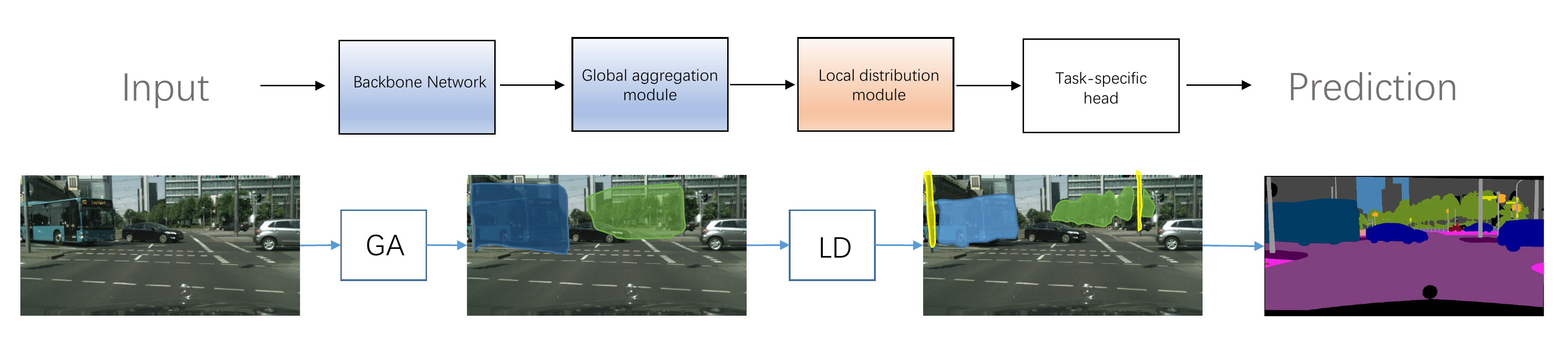}
\caption{
Our proposed GALD framework for semantic segmentation task. The imbalanced spread of information from small and large patterns in GA module is appropriately handled through LD module.
}
\label{fig:motivation}
\end{figure*}

In this paper, we propose a novel notion of {\em Local Distribution} (LD) module to tackle this problem. 
We introduce a LD module on top of 
any existing GA module
to distribute GA features adaptively (Figure~\ref{fig:motivation}).
The key idea is to condition GA features on position-wise pattern size in a way so that
local context information can be learned conditionally subject to global prior. 
Concretely, after the input features are processed by a GA module, we further learn local context with a LD module.
This collaboration between GA and LD
makes a new general architecture-agnostic GALD module.
Crucially, we design two different LD modules from two different aspects: sharpening pixel-wise GA features and exploring the relationships between GA features and LD features dynamically.

For the first aspect, we propose a simple yet effective module by estimating point-wised masks on GA features. Since GA modules calculate global statistics of features in large windows, they are easily
biased towards features from large patterns as they contain more samples. Then the global information distributed to each position is also biased towards large patterns, which causes over-smoothing results for small patterns. Started from the opposite view, if we sharpen the large objects, the small objects can be well preserved through residual-like design~\cite{resnet}. Intuitively, spatial operators for large patterns would shrink the spatial extent more while shrinking less even expand for small patterns. Thus we propose to use a set of downsampling operations followed by bilinear upsampling operation to capture the most salient components. After using sigmoid function point-wisely on the feature map, the large objects can be sharpened and the weight of large objects can be reduced. Meanwhile, the small objects can be well preserved. We name such LD module LDv1 which was published in BMVC2019~\cite{li2019global}.

For the second aspect, we propose a novel dynamic local distribution module according to the GA feature inputs. Motivated by recent self attention approaches~\cite{vaswani2017attention} and non-local-network~\cite{Nonlocal}, for each local position $p$, we calculate $k \times k$ different neighbor points affinity matrix based on both GA and LD features where the 
both inputs are considered together as the module inputs. In particular, we build the global-to-local graphs where the global context information is distributed to each local position dynamically. Different from previous non-local-like operators and their variants, we convert densely-connected graph with several local sparsely-connected graphs, which usually require much lower computational resources. Without loss of generality, for each location $p$, we propose to sample $k \times k$ region with $k=5$ by default where the affinity weights are calculated conditionally on the input GA features. The above neighbor sampling strategy will greatly reduce the complexities of both time and space from $\mathcal{O}(N^2)$ (for original non-local blocks~\cite{Nonlocal}) and $\mathcal{O}(N*( H + W-1))$ (for its variants CCNet~\cite{ccnet}) to $\mathcal{O}(N*k^2))$ where $k^2 \ll H $, $H$ and $W$ represent height and width of input resolution respectively, $k$ is kernel size of local sampling. 
We name such dynamic local distribution module LDv2 in this paper. {In particular, benefited from the flexible inference of  attention mechanism, our newly proposed outperform origin LDv2 by a large margin on various metrics(both mIoU and F1-score) on various baselines which can be found in experiment part.}

Both proposed LD modules can be combined with existing GA modules to form different GALD modules for various segmentation tasks. That means our work is orthogonal to all the previous work focusing on GA modeling.
In our experiment, LD is verified on GA modules such as PSP~\cite{pspnet}, ASPP~\cite{deeplabv3}, Non-Local~\cite{Nonlocal,ocnet} and CGNL~\cite{cgnl}, and achieves consistent performance improvement which are shown in the experiment part. In particular, a mean IoU score of 83.2\% on Cityscapes test set is achieved without using extra coarse annotated data with new proposed LDv2 modules. 
To the best of our knowledge, we are the \textit{first} to achieve 83\% mIoU on Cityscapes test set with \textbf{only fine annotation} and ResNet101 as backbone by the submission date of this paper.

A preliminary version of this work was published in~\cite{li2019global}.
In this paper we make the following significant 
extensions:
(i) We introduce a new and more effective local distribution module characterized by local attention to efficiently distribute global context.
Experiments on semantic segmentation show that this new design consistently outperforms our previous module.
(ii) We conduct comprehensive ablation studies to verify the proposed method including quantitative improvements over baselines and visualization analysis.
(iii) We conduct extensive experiments on four more challenging semantic segmentation datasets including Camvid~\cite{CamVid}, Pascal Context~\cite{pcontext}, ADE-20k~\cite{ADE20K} and COCO-stuff~\cite{coco_stuff} where our method establishes new state of the art.

%% file: 2relatedwork.tex
\section{Related work}
\label{sec:related}

\subsection{Semantic segmentation}
Often, all deep learning based semantic segmentation works to leverage the fully-convolutional networks (FCNs)~\cite{fcn}.
FCN removes the global aggregation layers such as global average pooling layer and fully-connected layers for semantic segmentation. The work~\cite{dilation} removed the last two downsample layers to obtain dense prediction and utilized dilated convolutions to enlarge the receptive field. 
Meanwhile, both SAC~\cite{sac} and DCN~\cite{deformable} improved the standard convolutional operator to handle the deformation and various scales of objects, which also enlarge the receptive fields of CNN operator.
Unet~\cite{unet}, SemanticFPN~\cite{upernet,PanopticFPN}, deeplabv3+~\cite{deeplabv3p}, SPGnet~\cite{cheng_spgnet}, RefineNet~\cite{refinenet} and DFN~\cite{dfn} adopted encoder-decoder structures that fuse information in low-level and high-level layers to make dense prediction results. Following such architecture design, GFFnet~\cite{xiangtl_gff}, CCLnet~\cite{ding2018context} and G-SCNN~\cite{gated-scnn} use gates for feature fusion to avoid noise and feature redundancy. CRF-RNN~\cite{crf_as_rnn} used graph models such CRF, MRF for semantic segmentation. AAF~\cite{aaf} proposed adversarial learning to capture and match the semantic relations between neighboring pixels in the label space. DenseDecoder~\cite{densedecoder} built multiple long-range skip connections on cascaded architecture. ICNet~\cite{ICnet},BiSegNet~\cite{bisenet} and SFNet~\cite{SFnet} were designed for real-time semantic segmentation. There are also specially designed video semantic segmentation works for boosting accuracy~\cite{Netwarp,GRFP_video} and saving inference time~\cite{DFF,Lowlatency_net}. DPC~\cite{DPC} and auto-deeplab~\cite{auto-deeplab} utilized architecture search techniques to build multi-scale architectures for semantic segmentation. Semantic segmentation is also actively studied in the topics of domain adaptation ~\cite{Tsai_adaptseg_2018}, model distillation ~\cite{Liu_distill_seg,He_knowled_adapt}, weakly supervised learning~\cite{Song_2019_CVPR,Wang_2020_CVPR}, semi-supervised learning~\cite{Chen-video-semi-seg}, few-shot learning~\cite{Zhang_2019_CVPR_canet} and interactive segmentation~\cite{ding2020phraseclick}.

\subsection{Global context aggregation}
To further enlarge the receptive field to the whole image, several methods are proposed recently. Global average pooled features are concatenated into existing feature maps in ~\cite{parsenet}. In PSPnet~\cite{pspnet}, average pooled features of multiple window sizes including global average pooling are upsampled to the same size and concatenated together to enrich global information. The DeepLab variants \cite{deeplabv1, deeplabv3,deeplabv3p} propose atrous or dilated convolutions and atrous spatial pyramid pooling (ASPP) to increase the effective receptive field. DenseASPP~\cite{denseaspp} improves on \cite{deeplabv2} by densely connecting convolutional layers with different dilation rates to further increase the receptive field of network. In addition to concatenating global information into feature maps, multiplying global information into feature maps also shows better performance~\cite{encodingnet, cbam, cgnl, dfn}. EncNet \cite{encodingnet} and SqueezeSeg \cite{SqueezeSeg} use attention along the channel dimension of the convolutional feature map to account for global context such as the co-occurrences of different classes in the scene. CBAM \cite{cbam} explores channel and spatial attention in a cascaded way to learn task specific representation. 

\begin{figure*}
	\centering
	\includegraphics[width=0.80\linewidth]{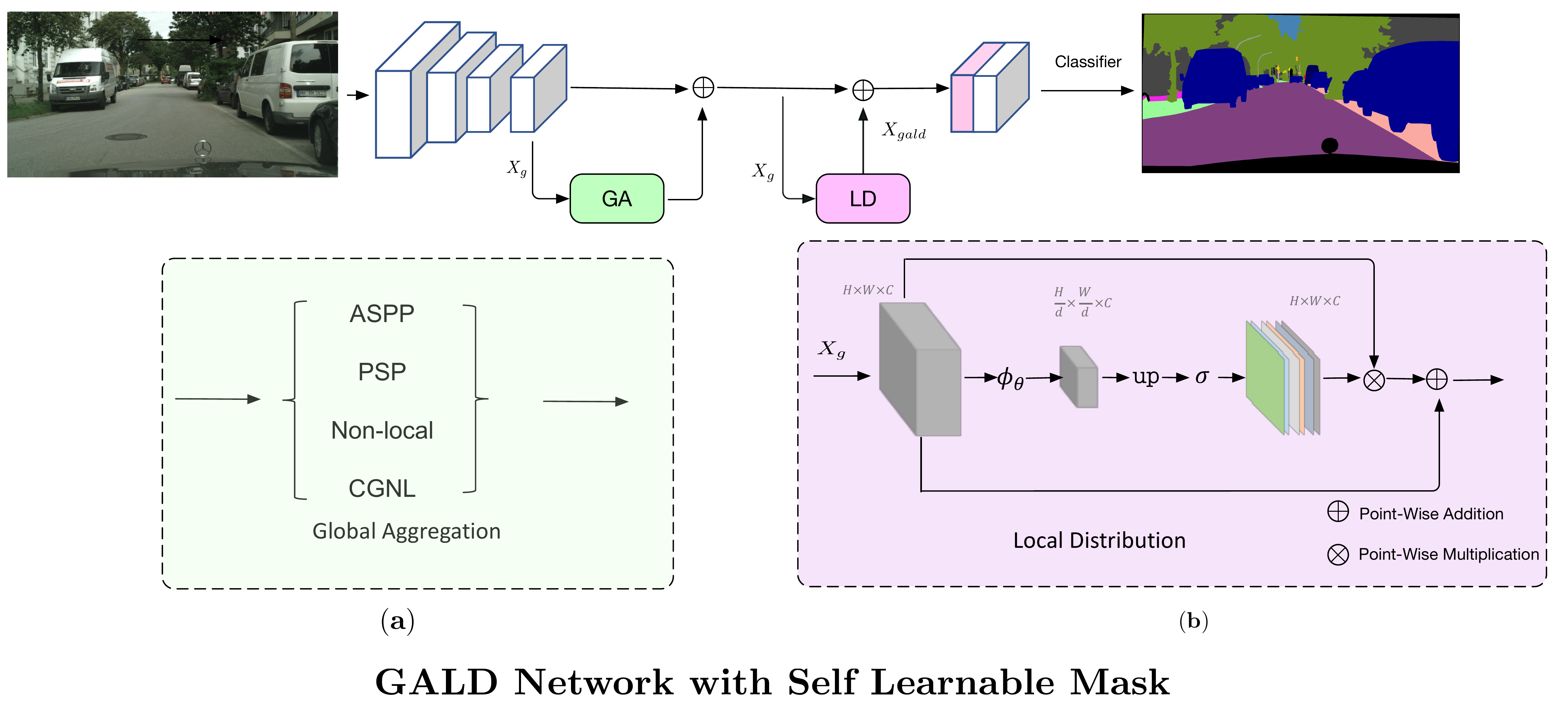}
	\caption{Schematic illustration of previous GALDv1~\cite{li2019global}, which contains two main components: Global Aggregation (GA) and Local Distribution (LD). 
		GALD receives a feature map from the backbone network and outputs a feature map with same size with global information appropriately assigned to each local position.
	}
	\label{fig:whole}
\end{figure*}

Recently, more advanced global information modeling approaches initiated from non-local network~\cite{Nonlocal} yield promising results on scene understanding tasks. In contrast to convolutional operator where information is aggregated locally defined by local filters, non-local operators aggregate information from the whole image based on an affinity matrix calculated among all positions around the image. Using non-local operator, impressive results are achieved in OCNet~\cite{ocnet},CoCurNet~\cite{CoCurrentNet}, DANet~\cite{DAnet}, A2Net~\cite{a2net}, CCNet~\cite{ccnet} and Compact Generalized Non-Local Net~\cite{cgnl}. OCNet~\cite{ocnet} uses non-local blocks to learn pixel-wise relationship while CoCurNet~\cite{CoCurrentNet} adds extra global average pooling path to learn whole scene statistic. DANet~\cite{DAnet} explores orthogonal relationships in both channel and spatial dimension using non-local operator. CCNet~\cite{ccnet} models the long range dependencies by considering its surrounding pixels on the criss-cross path through a recurrent way to save both computation and memory cost. Compact Generalized non-local Net~\cite{cgnl} considers channel information in affinity matrix. Another way to model the pixel-wised relationship is presented in PSANet~\cite{psanet}. It captures pixel-to-pixel relations using an attention module that takes the relative location of each pixel into account. EMANet~\cite{2019Expectation} adopts expectation-maximization algorithm~\cite{EMalgorithm} for the self-attention mechanism.

Another way to get global representation is using graph convolutional networks, and perform reasoning in a non-euclidean space~\cite{beyond_grids,chen2019graph,spg_net_gcn,DGM_net} where messages are exchanged within the graph space. 
Glore~\cite{chen2019graph} projects the feature map into interaction space using learned projection matrix and applies graph convolution on projected fully connected graph. BeyondGrids~\cite{beyond_grids} learns to cluster different graph nodes and uses graph convolution in parallel. SPGnet~\cite{spg_net_gcn} performs spatial pyramid graph reasoning while DGMnet~\cite{DGM_net} uses a dynamic graph reasoning framework for more efficient learning. Built on top of these global aggregation approaches, our work introduces orthogonal improvements.

\subsection{Local context exploration}
Most of the previous works on semantic segmentation focus on global contextual modeling.
Our work utilizes global information modeling but takes a further step to better distribute the global information to each local position.
There are a couple of works similarly focusing on local context modeling.
ExFuse~\cite{Exfuse} assigns auxiliary supervisions to the early stages of encoder for improving low-level context.
\cite{Fu_2019_ICCV} explores local context information via a gating operation for scene understanding. 
Our work is fundamentally different from these because,
(i) we adopt local affinity learning with global prior in a more adaptive manner, and
(ii) our LD module can be employed with various GA modules while previous works~\cite{Fu_2019_ICCV,xiangtl_gff,shuai2018toward} are limited to encoder-decoder framework with specific GA heads and hence less scalable to state-of-the-art architectures. 
 {Note that CGBNet~\cite{ding2020semantic} also explores the local global relationship via the difference of convolutions with various dilation rates,  which is different from our work that uses dynamic attention module guided by the global features. The SVCNet~\cite{SVCNet} proposes a shape-variant mask which shares the same spirit of self learnable mask in LDv1 module from the implementation aspect. However, the goal of their shape-variant mask is used to filter structure noise to model the global context while our self learnable mask is to better control and assign obtained global context, resulting in fine-grained feature representation. 
 We show that our approach is superior to ~\cite{Fu_2019_ICCV} in the experiment part.}

%% file: 3method.tex
\section{Method}

In this section, we first describe the GALD framework including the implementation details of LDv1. Then we give a more detailed description of the proposed LDv2.  

\subsection{GALD Model Overview}
Our GALD exploits long-range contextual information of input feature tensor $ \mathbf{X} \in \mathbb{R}^{ H \times W \times C }$  in a fully-convolution network (FCN), and adaptively distributes global context to each spatial and channel position of the output feature, $\mathbf{X}_{gald} \in \mathbb{R}^{ H \times W \times C }$.

\noindent{\textbf{Global Aggregation.}} 
To obtain global context at each position, GA takes feature vectors of $\mathbf{X}$ as input and learns features in a larger receptive field (up to the whole feature map size). 
Take the Compact Generalized Non-Local (CGNL)~\cite{cgnl} as an example, similar to non-local~\cite{Nonlocal}, it aggregates contextual information from all spatial and channel positions in the same group. 
Specifically, the global statistics are calculated for each group and multiplied back to the features in the same group, outputting a refined feature tensor $\mathbf{X}_{g}$. Note that, we downsample $\mathbf{X}$ by a factor of $2$ for saving memory and computation cost without sacrifice of performance.
This implies high redundancy of existing global aggregation. GALD is flexible in the selection of existing GA design.

We observe that global feature statistics are heavily biased towards large objects/patterns occupying more space/pixels. 
Consequently, when distributing global information evenly to each position, the representation of small patterns would be overwhelmed and become over-smoothing, leading to non-trivial performance drop.

\begin{figure*}
	\centering
	\includegraphics[width=0.80\linewidth]{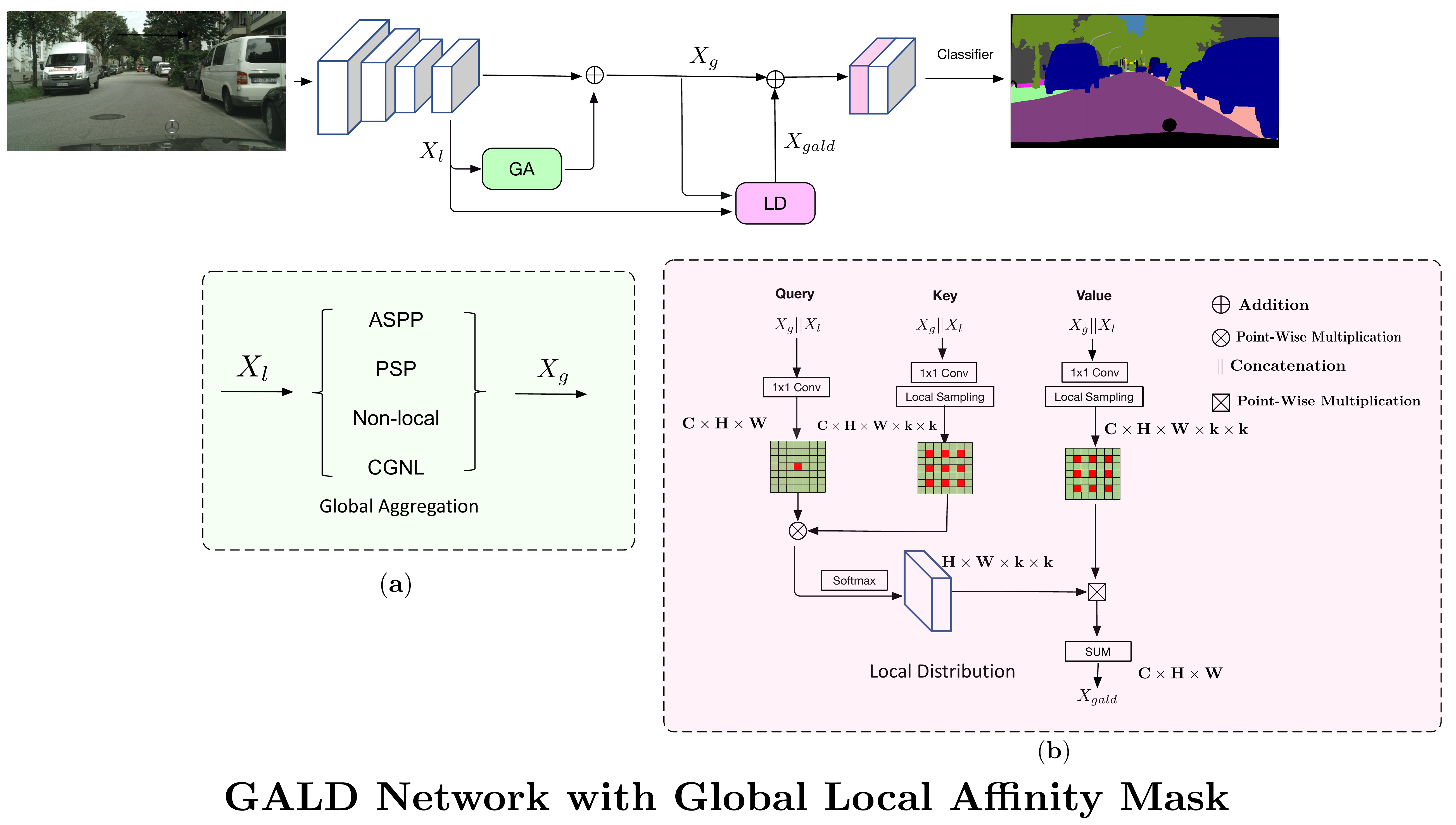}
	\caption{Schematic illustration of new prposed GALDv2, which contains two main components: Global Aggregation (GA) and Local Distribution (LD). GALD receives a feature map from the backbone network and outputs a feature map with same size with global information appropriately assigned to each local position. (a) means the global aggregation approaches. (b) means the proposed dynamic local distribution module.
	}
	\label{fig:whole_ldv2}
\end{figure*}

\noindent{\textbf{Local Distribution.}} 
LDv1 is proposed to adaptively use $\mathbf{X}_{g}$ considering patterns for each position. Without explicit supervision, the required patterns are latently described by $C$ channels in $\mathbf{X}_{g}$. For each pattern/channel $c\in\{1,...,C\}$, a spatial operator is learned to recalculate the spatial extent of the pattern in an image based on the activation map  $\mathbf{X}_{g}[:,:,c]$ sliced from $\mathbf{X}_{g}$. Intuitively, spatial operators for large patterns would shrink the spatial extent more while shrinking less even expand for small patterns.

The spatial operators for each channel is modeled as a stack of depth-wise convolutional layers with $\mathbf{X}_{g}$ as input:
\begin{align}
\mathbf{M} = \sigma(\text{upsample}(\mathbf W_{d} \mathbf{X}_{g})),
\end{align}
where $\mathbf{M} \in [0,1]^{H\times W \times C}$ contains the masks for each channel, accounting for 
refined spatial extent.
$\sigma(\cdot)$ is the sigmoid function, $\mathbf W_{d}$ is the weights of  those depth-wise convolutional filters with $d$ as the downsampling rate by stride convolution. The output mask $\mathbf{M}$ is sensitive to both spatial and channel and it is upsampled using bilinear interpolation. With $\mathbf{M}$, $\mathbf{X}_{g}$ is refined into $\mathbf{X}_{gald}$ in a residual manner by
\begin{align}
\mathbf{X}_{gald} = \mathbf{M} \odot \mathbf{X}_{g}  + \mathbf{X}_{g},
\end{align}
where $\odot$ is element-wise multiplication, and elements in $\mathbf{X}_{g}$ are weighted according to estimated spatial extent of each pattern at each position. In summary, LD predicts local weights $\mathbf{M}$ at each position of GA features for mitigating the over-smoothing effect of global aggregation.

Following~\cite{pspnet}, the original input feature $\mathbf{X}$ and global aggregated feature $\mathbf{F}_{GA}$ are concatenated together for final segmentation head, i.e.,
\begin{align}
\begin{split}
\mathbf{X}_{o} &= \text{concat}( \mathbf{X}_{gald}, \;\; \mathbf{X}) \\
&= \text{concat}( \mathbf{M} \odot \mathbf{X}_{g}  + \mathbf{X}_{g}, \;\; \mathbf{X} ),
\end{split}
\label{eq:final_feature}
\end{align}
where $\bf{M}$ adds point-wise trade-off between global information $\mathbf{X}_{g}$ and local detailed information $\bf{X}$ 
Conceptually, this module changes the proportion and distribution of coarse features in GA, yielding a fine-grained feature representation output $\mathbf{X}_{o}$.

\noindent {\bf Architecture.}
Figure~\ref{fig:whole} illustrates the overall architecture.
For semantic segmentation, 
GALD is placed on top of a backbone FCN.
The features from Eq.~\eqref{eq:final_feature} are used for final prediction.

\subsection{Dynamic Local Distribution Module}
Despite the simplicity and efficiency of LDv1, there are some existing problems in design. Firstly, LDv1 only considers the GA features and ignores the relationship between individual points of local features. It only refined the weights of GA features using learnable masks on the final representation for segmentation, whilst local features are not enhanced explicitly. Secondly, downsampling operation causes loss of high frequency components of GA features. This hurts the final performance on small parts of large objects like the window of a bus or truck. Thirdly, simple residual learning for local context can not guarantee the local pattern consistency in the scene since there is no explicit local context involved. 
To tackle these issues, we propose a novel dynamic local distribution module. Since our module is motivated by self-attention models and graph models, we will briefly describe them before giving the details of our approach.

\noindent \textbf{Review of Self-Attention Models}: We describe non-local network~\cite{Nonlocal} in view of a fully connected graphical model. For a 2D input feature with the size of $C \times H \times W$, where $C$, $H$, and $W$ denote the channel dimension, height, and width respectively, it can be interpreted as a set of features, $\ve{X} = [\ve{x}_1,\ve{x}_2, \cdots, \ve{x}_N]^\intercal$, $\ve{x}_i\in\mathbb{R}^C$, where $N$ is the number of nodes (\eg $N=H \times W$), and $C$ is the node feature dimension. 

\begin{equation}\label{eq:nonlocal-matrix}
\begin{aligned}
\ve{\tilde{X}} = \delta(\ve{X_\theta}\ve{X_\phi}^\intercal)\ve{X_g} 
=\ve{A}(\ve{X})\ve{X_g},
\end{aligned}
\end{equation}
where $\ve{A}(\ve{X}) \in \mathbb{R}^{N \times N}$ indicates the affinity matrix, $\ve{X_\theta} = \ve{W_\theta}\ve{X} \in \mathbb{R}^{N \times C'}$, $\ve{X_\phi} = \ve{W_\phi}\ve{X} \in \mathbb{R}^{N \times C'}$, and $\ve{X_g} = \ve{W_g}\ve{X} \in \mathbb{R}^{N \times C'}$ are obtained from the origin input $\ve{X}$ through $1\times1$ convolution for dimension reduction. $\ve{W_\theta}$, $\ve{W_\phi}$ and $\ve{W_g}$ mean the Query, Key and Value function respectively. The long range dependency can be captured by multiplying the global affinity matrix into $\ve{X_g}$

In this formulation (Equ.~\eqref{eq:nonlocal-matrix}) both the computation cost has a quadratic complexity with the number of node $N$. This indicates expensive computational cost
in semantic segmentation for natural images.

\noindent \textbf{Motivation.}
Our goal is to build a more efficient yet strong global-to-local relationship. Instead of propagating the messages from all the nodes exhaustively, we propagate global information into its relevant local positions adaptively. Consequently, the number of connected nodes for each anchor node can be reduced from $N$ to $K$ ($K \ll N$).
$K$ is defined as the number of local points: $K=k \times k$ where $k$ is the local kernel size.
This can dramatically reduce the prohibitive computation complexity. The total pipeline can be transformed as:

\begin{equation}
\begin{aligned}
\underbrace{{{\color{red}{X}}}^{{1} \times N} \times  {{\color{blue}{X}}}^{N \times C'}}_{=N} & \rightarrow
\underbrace{{\color{red}{X}}^{{1} \times {\color{red}{K}}} \times  {\color{blue}{X}}^{{\color{red}{K}} \times C'}}_{=N}, \\
\end{aligned}
\label{transform}
\end{equation}

Concretely, this reduces the computation cost from $\mathcal{O}(C' \times N^2)$ to $\mathcal{O}(C' \times N \times K)$.
For instance, given a $65 \times 65$ input feature, we have $N = 65 \times 65 = 4225$, v.s. $K = 25$ with $k=5$ in our default setting. Our method is hence more scalable than the non-local block.

\noindent \textbf{LDv2 Formulation:}
Under the aforementioned motivation, we construct the Global-to-Local graph by sampling neighbor nodes, as illustrated in Figure~\ref{fig:whole_ldv2} in the pink boxes.

Specifically, we start with concatenating GA $X_{g}$ and LD features $X_{l}$ and use it as the inputs $x$.
We denote the local sampled features as $s$. For each position $i$ of the node feature, we then sample a set of local neighbor node features from $x$ for Key and Value branches as following:
\begin{equation}
\label{eq:sample-func}
\mathcal{F}(\ve{x}(i)) = \left\{ \ve{s}(n)|n=1,2,\cdots, k \right\},
\end{equation}
where $s(n) \in \mathcal{R}^{C'}$ is the local node features sampled. In particular, we use the uniform sampling strategy with a dilation rate $r$ to sample the local features.
This is a local sampling process defined as:
\begin{equation}
\begin{aligned}
\mathcal{F}(\ve{x}(p)) &= \left\{ \ve{x}(p + \Delta r_n)| n=1,2,\cdots, k \right\}, 
\end{aligned}
\end{equation}
where $\Delta r_n$ is the $n$-th point with dilated rate $r$. In Figure~\ref{fig:whole_ldv2}, we choose $r=2$ for illustration. 

Given the sampling function $\mathcal{F}(\ve{x}_i)$, the entire process can be reformulated as:
\begin{equation}
\label{eq:global-local-graph}
\begin{aligned}
\ve{X_{gald}}(i) &= \sum_{\forall n} \delta(\ve{x}_{\theta}(i)\ve{s}_{\phi}(n)^\intercal) \ve{s}_g(n), \\
\end{aligned}
\end{equation}
where $n$ only enumerates the sampled positions, and $\delta$ is the softmax function. $\ve{x}_{\theta}(i) = W_{\theta}\ve{x}(i)$, $\ve{s}_{\phi}(i) = W_{\phi}\ve{s}(i)$, $\ve{s}_{g}(i) = W_{g}\ve{s}(i)$ refer to the Query, Key and Value functions respectively.
All the three functions are implemented by $1 \times 1$ convolution layer.

\noindent \textbf{Architecture.}
To have a more comprehensive representation, we combine refined feature $X_{gald}$ and original feature $X_l$ for final segmentation (Figure~\ref{fig:whole_ldv2} (b)). We use the dilated sampling strategy to handle various local context from diverse scenes. This avoids large and computationally expensive convolution kernels. Zero padding is adopted to handle the border cases. 

\noindent \textbf{Discussion on LDv2 module with related self attention module.}
{ For the LDv2 module, we model the global-to-local graph in an adaptive manner. Our approach is motivated by the self-attention network~\cite{vaswani2017attention}. However, it is \textbf{different} from the original design. First, our LDv2 is based on our original GALD framework~\cite{li2019global}. In particular, for each sampled local patch, we calculate the relationship to each point where the computation cost is significantly reduced and the global information is adaptively controlled by the affinities. We have proved the effectiveness of LDv2 over LDv1 in the experiment section where we achieve \textbf{over \% 1.5-2 mIoU} gain on various baselines. }

%% file: 4experiment.tex
\section{Experiment}

In this section, we evaluate our GALD framework on five scene parsing datasets. For both two LD modules, we carry out detailed analysis and ablation studies on Cityscapes dataset. We also report results on all datasets using the best settings found in ablation. Moreover, we give both detailed visual analysis on error maps and quantitative comparison on object segmentation boundaries. Finally we explore our methods on remaining semantic segmentation datasets.

\subsection{Benchmarks} 

\noindent
\textbf{Cityscapes:} Cityscapes ~\cite{Cityscapes} is a benchmark densely annotated for 19 categories in urban scenes, which contains 5000 fine annotated images in total and is divided into 2975, 500, and 1525 images for training, validation and testing, respectively. In addition, 20,000 coarse labeled images are also provided to enrich the training data. Images are all with the same high resolution in the road driving scene, i.e., $1024 \times 2048$. Note that, we use fine dataset for ablation study and comparison with previous methods. We also use the coarse data for fair comparison on the benchmark.

\noindent
\textbf{Pascal Context:} Pascal Context~\cite{pcontext} provides pixel-wise segmentation annotation for 59 classes and 1 background class. The training and testing sets consist of about 4998 and 5105 images respectively.

\noindent
\textbf{ADE-20k:} ADE-20k~\cite{ADE20K} is a challenging dataset that contains 22K densely annotated images with 150 fine-grained semantic concepts. The training and validation sets consist of 20210 and 2000 images respectively. Both pixel accuracy and mIoU are reported. 

\noindent
\textbf{COCO-stuff:} COCO-stuff~\cite{coco_stuff} has 171 categories including 80 objects and 91 stuff annotated to each pixel. Following previous works~\cite{SpyGR} we adopt 9,000 images for training and 1,000 images for testing. We report mIoU as the evaluation metric.

\noindent
\textbf{Camvid:} Camvid~\cite{CamVid} is another road scene dataset. This dataset involves 367 training images, 101 validation images and 233 testing images with resolution of $960 \times 720$.

\begin{figure*}
	\centering
	\includegraphics[width=0.99\linewidth]{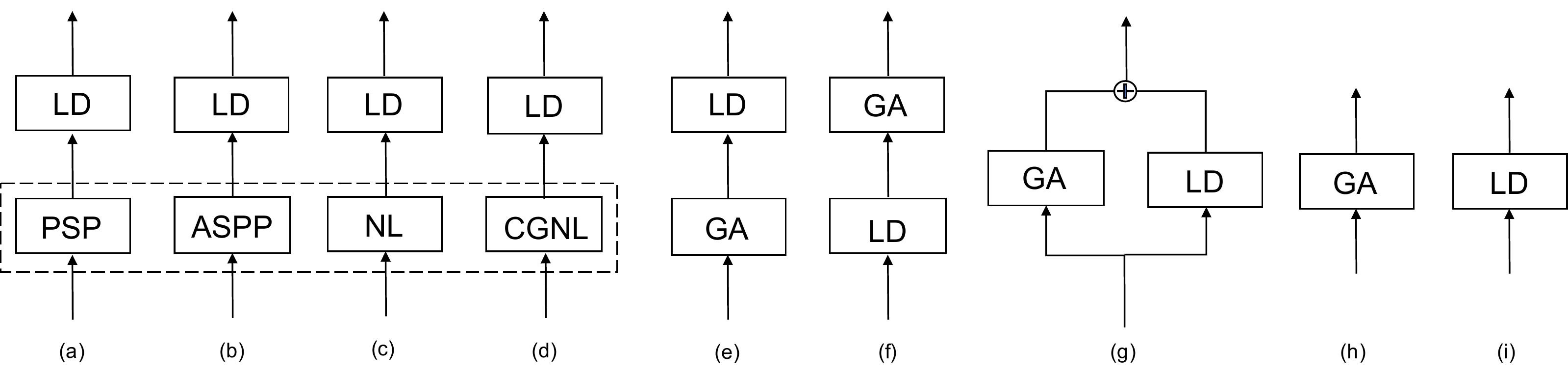}
	\caption{Ablation studies on GA and LD and their combinations. (a-d): different GA modules with LD. (e-g): different arrangements of GA and LD. (h-i): using GA and LD individually.}
	\label{fig:ablation}
\end{figure*}

\subsection{Implementation Details}

\noindent{\bfseries Semantic Segmentation on Cityscapes} 
Following the same setting as PSPNet~\cite{pspnet},
we employ Fully Convolutional Networks (FCNs) as baseline with ImageNet pretrained ResNet as the backbone. The proposed GALD is appended to the backbone with random initialization. For optimization, we also keep the same training setting as PSPNet, where mini-batch SGD with momentum 0.9 and initial learning rate 0.01 is used to train all models with 50K iterations, using mini-batch size of 8 and crop size of 832 for Cityscapes. During training, ``poly'' learning rate scheduling policy with power = 0.9 is used to adjust the learning rate. Synchronized batch normalization~\cite{pspnet} is used for fair comparison with previous works. We will give detailed settings further for the remaining datasets.
To further boost the performance, Online Hard Example Mining (OHEM) loss~\cite{ohem} is used for training, where only top-K ranked pixels in terms of loss are used during back-propagation. Same as the PSPNet~\cite{pspnet}, we also add auxiliary loss at the stage-4 of ResNet for more stable training.

\begin{table*}[!t]
	\centering	
	\caption{Comparison results on Cityscapes validation set, where $\Delta a$ denotes the performance difference comparing with baseline, and $\Delta b$ denotes performance difference between using GALD module and the corresponding GA module.
		All methods are evaluated with the single-scale crop test using ResNet50 as backbone.}		
	\begin{minipage}{\dimexpr.40\linewidth}
		\centering
		\tiny
		\resizebox{0.75\textwidth}{!}{%
			\begin{tabular}{ l|c|c }
				\toprule[0.2em]
				Method & mIoU(\%) & $\Delta a$(\%) \\
				\toprule[0.2em]
				FCN (Baseline) & 73.7  & - \\
				\hline\hline
				+ASPP~\cite{deeplabv3} & 77.2 & 3.5 $\uparrow$ \\ 
				
				+NL~\cite{Nonlocal} & 78.0 & 4.3 $\uparrow$ \\
				+PSP~\cite{pspnet} &  76.2 & 2.5 $\uparrow$ \\
				+CGNL~\cite{cgnl} & \textbf{78.2} & \textbf{4.5} $\uparrow$ \\
				\hline
			\end{tabular}
		}
		\par
		{\footnotesize(a) Ablation study on different GA modules.}
	\end{minipage}
	\begin{minipage}{\dimexpr.50 \linewidth}
		\centering
		\tiny
		\resizebox{0.75\textwidth}{!}{%
			\begin{tabular}{ l|c|c|c }
				\toprule[0.2em]
				Method & mIoU(\%) & $\Delta a$(\%) & $\Delta b$(\%)\\
				\toprule[0.2em]
				FCN (Baseline) & 73.7  & - & -\\
				\hline
				\hline
				+LDv1 & 77.5 & 3.8 $\uparrow$ & - \\
				+PSP + LDv1 & 78.9 & 5.2 $\uparrow$ & \textbf{2.7} $\uparrow$ \\ 
				+ASPP + LDv1 & 79.5 & 5.4 $\uparrow$ & 2.3 $\uparrow$ \\
				+NL + LDv1 & 79.2 & 5.3 $\uparrow$ & 1.2 $\uparrow$ \\
				+CGNL + LDv1 & \textbf{79.6} & \textbf{5.9 $\uparrow$} & 1.4 $\uparrow$ \\
				\hline
			\end{tabular}
		}\par
		{\footnotesize(b) Ablation study on LDv1 applied on different GA modules.}
		
	\end{minipage}
	\vspace{3mm}
	\begin{minipage}{\dimexpr.40 \linewidth}
		\centering
		\tiny
		\resizebox{0.75\textwidth}{!}{%
			\begin{tabular}{ l|c|c}
				\toprule[0.2em]
				Method & mIoU(\%) & $\Delta a$(\%)\\
				\toprule[0.2em]
				FCN (Baseline) & 73.7  & - \\
				\hline
				\hline
				+Parallel & 77.5 & 3.8 $\uparrow$\\ 
				+LDGA  & 78.1 & 4.4 $\uparrow$ \\
				+GALD  & \textbf{79.6} & \textbf{5.9 $\uparrow$}\\
				\hline
			\end{tabular}
		}\par
		{\footnotesize(c) Ablation study on different arrangements of GA and LD .}
		
	\end{minipage}	
	\begin{minipage}{\dimexpr.520 \linewidth}
		\centering
		\tiny
		\resizebox{0.75\textwidth}{!}{%
			\begin{tabular}{ l|c|c|c }
				\toprule[0.2em]
				Method & mIoU(\%) & $\Delta a$(\%) & $\Delta b$(\%)\\
				\toprule[0.2em]
				FCN (Baseline) & 73.7  & - & -\\
				\hline
				\hline
				+LDv2 & 78.9 & 3.8 $\uparrow$ & - \\
				+PSP + LDv2 & 80.2 & 5.2 $\uparrow$ & \textbf{2.7} $\uparrow$ \\ 
				+ASPP + LDv2 & \textbf{80.8} & 5.4 $\uparrow$ & 2.3 $\uparrow$ \\
				+NL + LDv2 & 79.9 & 5.3 $\uparrow$ & 1.2 $\uparrow$ \\
				+CGNL + LDv2 & 80.1 & \textbf{5.9 $\uparrow$} & 1.4 $\uparrow$ \\
				\hline
			\end{tabular}
		}\par
		{\footnotesize(d) Ablation study on LDv2 applied on different GA modules.}
		
	\end{minipage}
	\vspace{1mm}
	\label{tab:city_ablation}
\end{table*}

\subsection{Ablation Study and Analysis}

\subsubsection{GALD framework Design}
\label{sec:ablation_on gald_framework}
\begin{figure}
	\centering
	\includegraphics[width=1.0\linewidth]{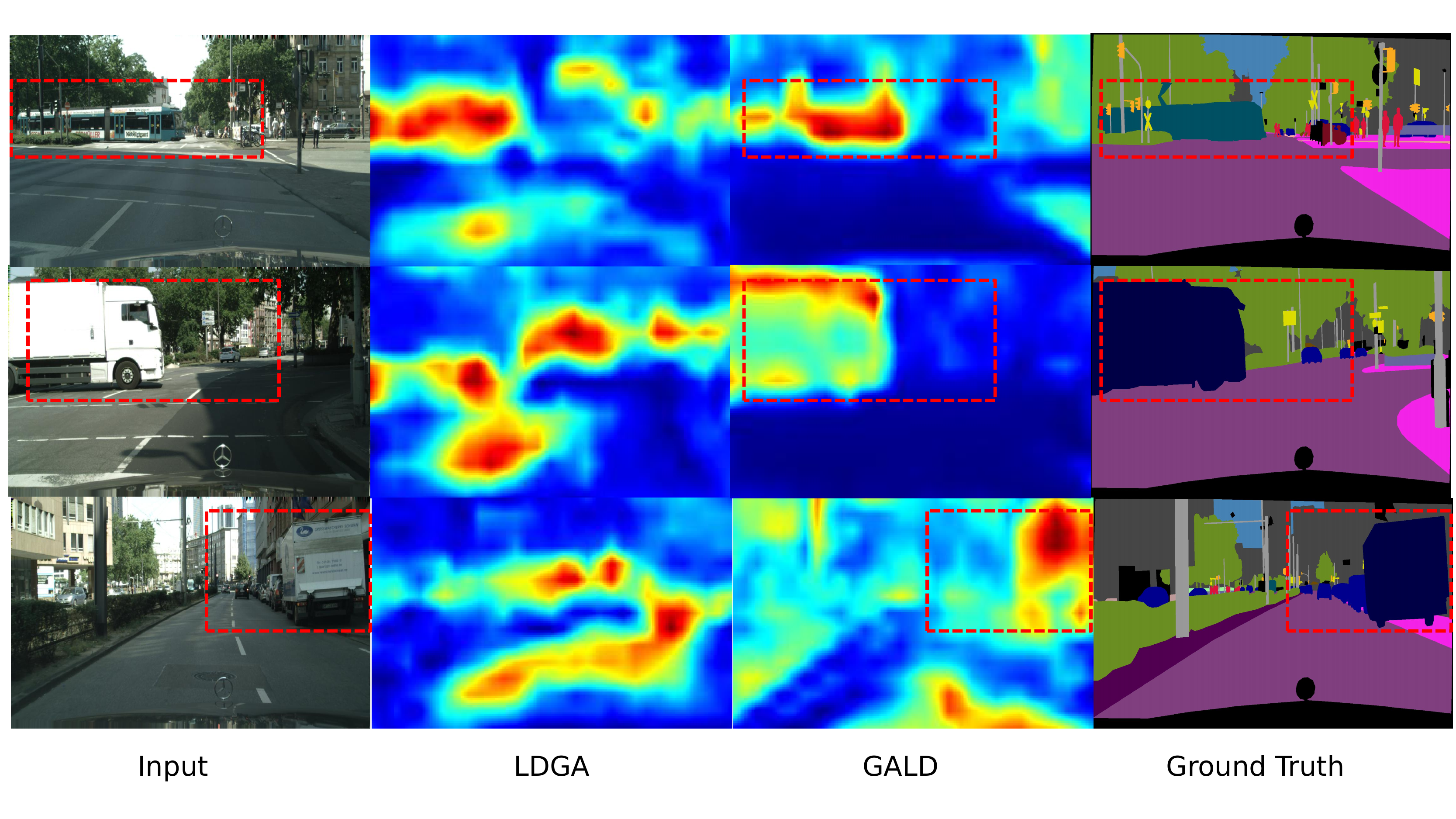}
	\caption{
		Comparison of mask maps learned in different arrangements of GA and LD. The mask maps are calculated by the mean of $\textbf{M}$ along channel dimension.
		Best viewed in color.}
	\label{fig:attenion_mask_2}
\end{figure}

{There are many choices for arrangement of both GA and LD. Thus, we first perform experiments to find the best setting for both GA and LD.}
We first explore our GALD framework design using LDv1 with four different GA modules as illustrated in Figure~\ref{fig:ablation} (a)-(c). Table~\ref{tab:city_ablation}(a) first reports the performances of adding four GA modules to the baseline FCN, where all methods are using the same backbone ResNet50 for fair comparison. Obviously, all GA modules significantly improve the baseline FCN on semantic segmentation task, where CGNL performs better than the other three GA modules. {This indicates the non-local-like operator has better performance than PPM and ASPP, which is also verified by a lot of recent works~\cite{OCRnet,annet}.}
Table~\ref{tab:city_ablation}(b) reports the results by adding our proposed LD module. Directly using LD alone improves the baseline FCN by 3.8\%, which demonstrates that features from FCN have the similar problem as features from GA modules. LD together with four different GA modules consistently improves the corresponding GA module. Comparing with baseline, the combination of CGNL+LD achieves the best performance. We choose CGNL as our default GA module in the following experiments for LDv1 module. Considering LD module can also improve the baseline, we further study different arrangements of LD and GA as illustrated in Figure~\ref{fig:ablation} (e)-(g). (f) and (g) represent LDGA and parallel in Table~\ref{tab:city_ablation}(c) respectively. LDGA means first-LD-then-GA while parallel means concatenating the output of LD and GA. Table~\ref{tab:city_ablation}(c) reports the results of the three different arrangements. All improve the baseline and GALD achieves best result. Figure~\ref{fig:attenion_mask_2} shows the mask maps learned in LDGA and GALD. It is shown that mask maps learned by GALD focus more on regions inside large objects and stress more the corresponding global features. 
In contrast, mask maps from LDGA have no obvious focus on large objects since the LD module has not accessed to global feature yet. For our new proposed LDv2, we carry out the same experiment as LDv1. As shown in Table~\ref{tab:city_ablation}(d), LDv2 yields more performance gain than LDv1. In particular, with ASPP as GA heads, our LDv2 can obtain 7.1 point gain on the baseline model and 3.6 point gain on GA head baseline. {The CGNL is lower than ASPP, which is mainly because the proposed LDv2 shares the similar attention mechanism from CGNL design. Moreover, ASPP is more general and fair to compare with previous work~\cite{gated-scnn}.} Thus we choose ASPP as our GA module for the following experiments for LDv2 module.

\begin{table}[!t]		
	\centering
	\caption{Comparison results on Cityscapes validation set, where $\Delta a$ denotes the performance difference comparing with baseline, and $\Delta b$ denotes performance difference between using GALD module and the corresponding GA module.
		All methods are evaluated with the single-scale crop test.}
	\label{tab:city_ablation_component_design}	
	
	\begin{minipage}{\dimexpr 0.8\linewidth}
		\centering
		\tiny
		\resizebox{1.0\textwidth}{!}{%
			\begin{tabular}{ l|c|c}
				\toprule[0.2em]
				Method & mIoU(\%)  & $\Delta b$ (\%)\\
				\toprule[0.2em]
				FCN (Baseline) & 73.7  & -\\
				FCN + CGNL & 78.2 & - \\
				\hline
				\hline 
				+CGNL+LD(depth-wise convolution)  & \textbf{79.6} & \textbf{1.4 $\uparrow$}\\
				+CGNL+LD(bilinear interpolation) & 77.6  & 0.6 $\downarrow$ \\
				+CGNL+LD(average pooling) & 76.5  & 1.7 $\downarrow$ \\
				\hline
			\end{tabular}
		}\par
		\vspace{1mm}
		{\footnotesize(a) Ablation study on downsampling strategies for mask estimation in LDv1 using ResNet50 as backbone, where the downsamping ratio is 8.}
	\end{minipage}
	\vspace{2mm}
	
	\begin{minipage}{\dimexpr.8 \linewidth}
		
		\tiny
		\resizebox{0.85\textwidth}{!}{%
			\begin{tabular}{ l|c|c}
				\toprule[0.2em]
				Method & mIoU(\%)  & $\Delta b$ (\%) \\
				\toprule[0.2em]
				FCN (Baseline) & 73.7  & -\\
				+LDv2($k=3\times3l r=3$) & 76.2 & 2.5$\uparrow$\\
				+LDV2($k=5\times5, r=3$) & 77.2 & 3.5$\uparrow$\\
				+LDV2($k=5\times5, r=6$) & 77.5 & 3.8$\uparrow$\\
				+LDV2($k=7\times7, r=6$) & 78.9 & 5.2$\uparrow$\\
				\hline
				\hline 
				FCN + ASPP(Baseline) & 77.2 & - \\
				+ASPP+LDv2($k=3\times3, r=3$) & 79.9 & 2.7$\uparrow$ \\
				+ASPP+LDv2($k=3\times3, r=6$) & 79.8 & 2.6$\uparrow$ \\
				+ASPP+LDv2($k=5\times5, r=3$) & 80.2  & 3.0$\uparrow$ \\
				+ASPP+LDv2($k=5\times5, r=6$) & 80.8 & 3.8$\uparrow$ \\
				+ASPP+LDv2($k=7\times7, r=6$) & 80.3  &  3.1$\uparrow$ \\
				\hline
			\end{tabular}
		}\par
		\vspace{1mm}
		{\footnotesize(b) Ablation study on local kernel size in LDv2 using ResNet50 as backbone.}
		\vspace{2mm}
	\end{minipage}
	
	\begin{minipage}{\dimexpr.8 \linewidth}
		
		\tiny
		\resizebox{0.85\textwidth}{!}{%
			\begin{tabular}{c c c l l}
				\toprule[0.2em]
				Method &  $X_{g}$ & $X_{l}||X_{g}$ & mIoU(\%) & Time \\  
				\toprule[0.2em]
				Key \& Query(baseline) & - & \checkmark &  80.8  \\
				Key & \checkmark & - & 80.1 & 0.7 $\downarrow$ \\
				Query & \checkmark  & - & 79.8 & 1.0 $\downarrow$ \\
				\midrule
			\end{tabular}
		}\par
		\vspace{1mm}
		{\footnotesize(c) Ablation study on Key/Query inputs in LDv2 using ResNet50 as backbone.}
		
	\end{minipage}

\end{table}

\begin{table}[!t]
	\centering
	\caption{Ablation study on improvement strategy on the validation set. MG means we use multi-grid strategy in the ResNet backbone. MS means multi scale test with flipping.}
	\label{tab:city_improvement}
	\begin{minipage}{\dimexpr.85\linewidth}
		\centering
		\resizebox{0.75\textwidth}{!}{%
			\begin{tabular}{c c c c l l }
				\toprule[0.2em]
				Method & ResNet101 & MG & MS & mIoU(\%) & $\Delta a(\%) $  \\  
				\toprule[0.2em]
				&  &  &  & 80.8 & -  \\ 
				& \checkmark & & & 81.5 & 0.3 $\uparrow$ \\
				& \checkmark & \checkmark & & 81.8 & 1.0 $\uparrow$ \\
				& \checkmark & \checkmark & \checkmark & 83.0 & 2.2 $\uparrow$ \\
				
				\hline
			\end{tabular}
		}\par
		\vspace{1mm}
	\end{minipage}
	
\end{table}

\begin{table}[!t]
	\centering
	\caption{Computaion Time Comparison with an input image of size $1 \times 1 \times 512 \times 512$ where we use ASPP module as our GA head.}
	\label{tab:efficiency}
	\begin{minipage}{\dimexpr.95\linewidth}
		\centering
		\resizebox{0.95\textwidth}{!}{%
			\begin{tabular}{c c c c l l l}
				\toprule[0.2em]
				Method & GA (ASPP) & LDv1 & LDv2 & mIoU(\%) & Time & {Param(M)} \\  
				\toprule[0.2em]
				FCN (Baseline) & - & - & - & 73.7 & 60ms & - \\ 
				& \checkmark & - & - & 77.2 & 65ms &  {47.5M} \\
				& \checkmark & \checkmark & - & 79.5 & 70ms & {+0.4M}  \\
				& \checkmark & - & \checkmark & 80.8 & 82ms & {+2.3M} \\
				FCN (ResNet101) &  \checkmark & -& - & 79.3 & 120ms & {65.4M} \\
				\midrule
			\end{tabular}
		}\par
		\vspace{1mm}
		{ Note that we use ASPP as our GA head. The FPS is calculated with the average of 1000 input images on a single V100 card.}	
	\end{minipage}
	
\end{table}

\subsubsection{LD component design}
\label{sec:ablation_on gald_component_design} 
{This section we perform experiments on LD component design including both LDv1 and LDv2 since there may be many design choices for each LD.} We first explore three different downsampling strategies for LD, including average pooling, bilinear interpolation and depth-wise stride convolution. Table~\ref{tab:city_ablation_component_design}(a) shows that depth-wise stride convolution achieves the best result, while average pooling and bilinear interpolation even slightly degrade the performance.  
{This indicates that the learnable filters for each channel are important to refine the features from the GA module. Learnable downsampling extracts most salient global features related to each local parts.} Then we give a detailed design on LDv2. {We mainly consider the kernel size of LDv2 since it controls the number of the sampled points.} Table~\ref{tab:city_ablation_component_design}(b) shows the performance effect of kernel size $k$ and dilation rate $r$ for both backbone only and backbone with ASPP as GA head. As shown in the top part of the table, increasing the kernel size and dilation rate can lead to consistent performance gain over backbone. Our LDv2 even further improves over GA heads with ASPP.
{As evident in the bottom half of the table, with the increase of kernel size and dilation rate, the performance gain is mostly increasing but with more parameters and computational cost. Thus we choose the proper setting by considering both effectiveness and efficiency.} The best setting with kernel size $k=5$ and dilation rate $r=3$ is used for the remaining datasets.

\begin{figure*}
	\centering
	\includegraphics[width=0.90\linewidth]{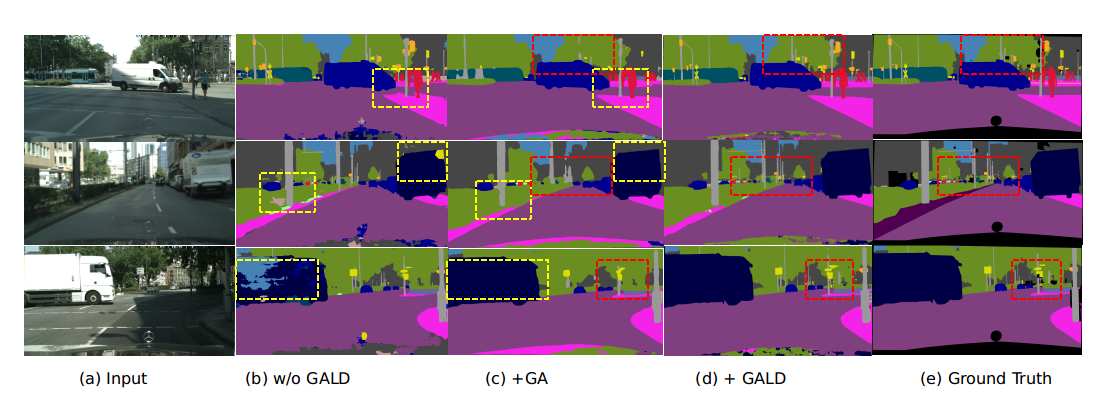}
	\caption{
		Visualization of different parts output results in one model.(a), input images; (b),results after FCN's outputs; (c), results after GA module's outputs; (d), results after GALD module'ss outputs;
		(e), ground truth. Yellow boxes highlight regions that GA can handle global semantic consistency, while red boxes highlight regions that LD can recover more detailed information.
		(Best viewed in color).}
	\label{fig:gald_mechanism}
\end{figure*}

\begin{table*}[t!]
	\centering 
	\small
	\caption{ 
		\small Per-category F-score results on the Cityscapes val set for 4 different thresholds. Note that our methods output both Deeplabv3+~\cite{deeplabv3p} and G-SCNN~\cite{gated-scnn} over \textbf{all four thresholds}. Best view on screen and zoom in.
	} 
	\addtolength{\tabcolsep}{0pt}
	\resizebox{\textwidth}{!}{
		\begin{tabular}{l|c|c|ccccccccccccccccccc}
			\toprule[0.2em]
			Method & Thrs & F1-score &  road & swalk & build. & wall & fence & pole & tlight & sign & veg & terrain & sky & person & rider & car & truck & bus & train & motor & bike \\
			\toprule[0.2em]
			Deeplabv3+~\cite{deeplabv3p} & 3px & 69.7 & 83.7 & 65.1 & 69.7 & 52.2 & 46.2 &  72.2 & 62.8 & 67.7 & 71.8 & 52.2  & 80.9  & 61.5 & 66.4 & 78.8 & 78.2 & 83.9 & 91.7 & 77.9 & 60.9 \\ 
			G-SCNN~\cite{gated-scnn} & 3px & 73.6 & 85.0 & 68.8 &  74.1 & 53.3 & 47.0 &  79.6 & 74.3 & \bf{76.2} & 75.3 & 53.1  & 83.5  & 69.8 & 73.1 & 83.4 & 75.8 & 88.0 & \bf{93.9} & 75.1 & 68.5 \\ 
			Ours (LDv1) & 3px & 73.7 & 85.1 & 67.8 &  \bf{75.1} & \bf{53.3} & 47.2 &  \bf{80.6} & 74.0 & 75.9 & \bf{75.6} & 53.8  & \bf{84.5}  & 68.9 & \bf{73.2} & \bf{83.5} & \bf{76.8} & 88.1 & 92.8 & 75.0 & 67.5 \\ 
			Ours (LDv2) & 3px & \bf{74.0} & \bf{85.2} & \bf{69.1} &  74.0 & 50.3 & \bf{50.2} &  79.6 & \bf{74.6} & 75.2 & 75.1 & \bf{55.3}  & 81.5  & \bf{70.2} & 72.1 & 82.4 & 76.3 & \bf{89.1} & 92.8 & \bf{76.2} & \bf{70.5} \\ 
			\hline
			Deeplabv3+~\cite{deeplabv3p} & 5px & 74.7 & 88.1 & 72.6 & 78.1 & 55.0 & 49.1 & 77.9 &
			69.0 & 74.7 & 81.0 & 55.8 & 86.4 & 69.0 &
			71.9 & 85.4 & 79.4 & 85.4 & 92.1 & 79.4 & 68.4 \\  
			G-SCNN~\cite{gated-scnn} & 5px & {77.6} &{88.7} & 75.3 & 80.9 & {55.9} & {49.9} & \bf{83.6} & {78.6} & {80.4} & \bf{83.4} & {56.6} & {88.4} & {75.4} &
			{77.8} & {88.3} & 77.0 & {88.9} & {94.2} & 76.9 & {75.1} \\\
			Ours (LDv1) & 5px & {79.4} & 88.6 & 74.6 &  \bf{81.8} & 55.2 & 55.3 &  83.3 & \bf{80.0} & \bf{80.6} & 82.9 & 60.3  & 88.2  & 75.4 & \bf{79.5} & 89.2 & 83.6 & \bf{92.8} & 96.3 & 80.9 & \bf{75.5} \\
			Ours (LDv2) & 5px & \bf{79.7} & \bf{89.4} & \bf{76.0} &  81.2 & \bf{66.1} & \bf{60.3} &  80.9 & 78.5 & 80.2 & 82.6 & \bf{61.8}  & \bf{88.4}  & \bf{76.0} & 78.2 & \bf{89.4} & \bf{84.0} & 92.3 & \bf{96.6} & \bf{82.3} & 75.1 \\
			\hline
			Deeplabv3+~\cite{deeplabv3p} & 9px & 78.7 & 91.2 & 78.3 & 84.8 & 58.1 & 52.4 & 82.1 &
			73.7 & 79.5 & 87.9 & 59.4 & 89.5 & 74.7 &
			76.8 & 90.0 & {80.5} & 86.6 & 92.5 & {81.0} & 75.4 \\ 
			G-SCNN~\cite{gated-scnn} & 9px & {80.7} & {91.3} & {80.1} & {86.0} & {58.5} & {52.9} & \bf{86.1} &
			{81.5} & {83.3} & {89.0} & {59.8} & {91.1} & {79.1} &
			{81.5} & {91.5} & 78.1 & {89.7} & {94.4} & 78.5 & {80.4} \\
			Ours (LDv1) & 9px & {82.4} & {91.5} & 79.7 & \bf{87.4} & 57.7 & 58.3 & 86.1 & \bf{83.1} & 83.8 & 88.9 & 63.7 & 90.8 & 79.3 & \bf{83.5} & 92.5 & {84.6} & \bf{93.5} & 96.6 & {82.4} & \bf{82.4} \\
			Ours (LDv2) & 9px & \bf{83.2} & \bf{92.2} & \bf{81.2} & 87.3 & \bf{63.7} &\bf{63.4} & 84.3 & 81.8 & \bf{83.9} & \bf{89.2} & \bf{65.1} & \bf{91.5} & \bf{80.7} & 82.5 & \bf{93.6} & \bf{85.1} & 93.1 & \bf{96.8} & \bf{84.1} & 81.2 \\
			\hline
			Deeplabv3+~\cite{deeplabv3p} & 12px & 80.1 & {92.3} & 80.4 & 87.2 & 59.6 & 53.7 & 83.8 &
			75.2 & 81.2 & 90.2 & 60.8 & 90.4 & 76.6 &
			78.7 & 91.6 & {81.0} & 87.1 & 92.6 & {81.8} & 78.0 \\ 
			G-SCNN~\cite{gated-scnn} & 12px & {81.8} & 92.2 & {81.7} & {87.9} & {59.6} & {54.3} & {87.1} &
			{82.3} & {84.4} & {90.9} & {61.1} & {91.9} & {80.4} &
			{82.8} & {92.6} & 78.5 & {90.0} & {94.6} & 79.1 & {82.2} \\
			Ours (LDv1) & 12px & {83.5} & {92.4} & {81.5} & {89.4} & {58.8} & 59.5 & \bf{87.1} & \bf{83.9} & 84.9 & 91.0 & {65.0} & 91.6 & {80.6} & \bf{84.9} & {93.5} & 85.1 & 93.7 & {96.7} & 82.9 & 83.1 \\
			Ours (LDv2) & 12px & \bf{84.3} & \bf{93.1} & \bf{82.9} & \bf{89.4} & \bf{64.8} & \bf{64.6} & 85.5 & 82.7 & \bf{85.2} & \bf{91.3} & \bf{66.3} & \bf{92.4} & \bf{82.1} & {83.9} & \bf{94.2} & \bf{85.5} & \bf{93.4} & \bf{97.0} & \bf{84.7} & \bf{83.2} \\
			\hline
		\end{tabular}
	}
	\vspace{2mm}
	\label{tab:cityscapes_results_detail_f_score}
\end{table*}

\subsubsection{Training strategies}
In comparison with other state-of-the-art models, we further adopt several common strategies to improve the performance of our model. Switching with ResNet-101 backbone gets 0.7\% mIoU gain while using multi-grid strategy in backbone leads to 0.3\% mIoU gain. After adopting multi-scale inference with flip, our final model obtains 83.0\% on validation dataset.

\begin{figure}
	\centering
	\includegraphics[width=0.65\linewidth]{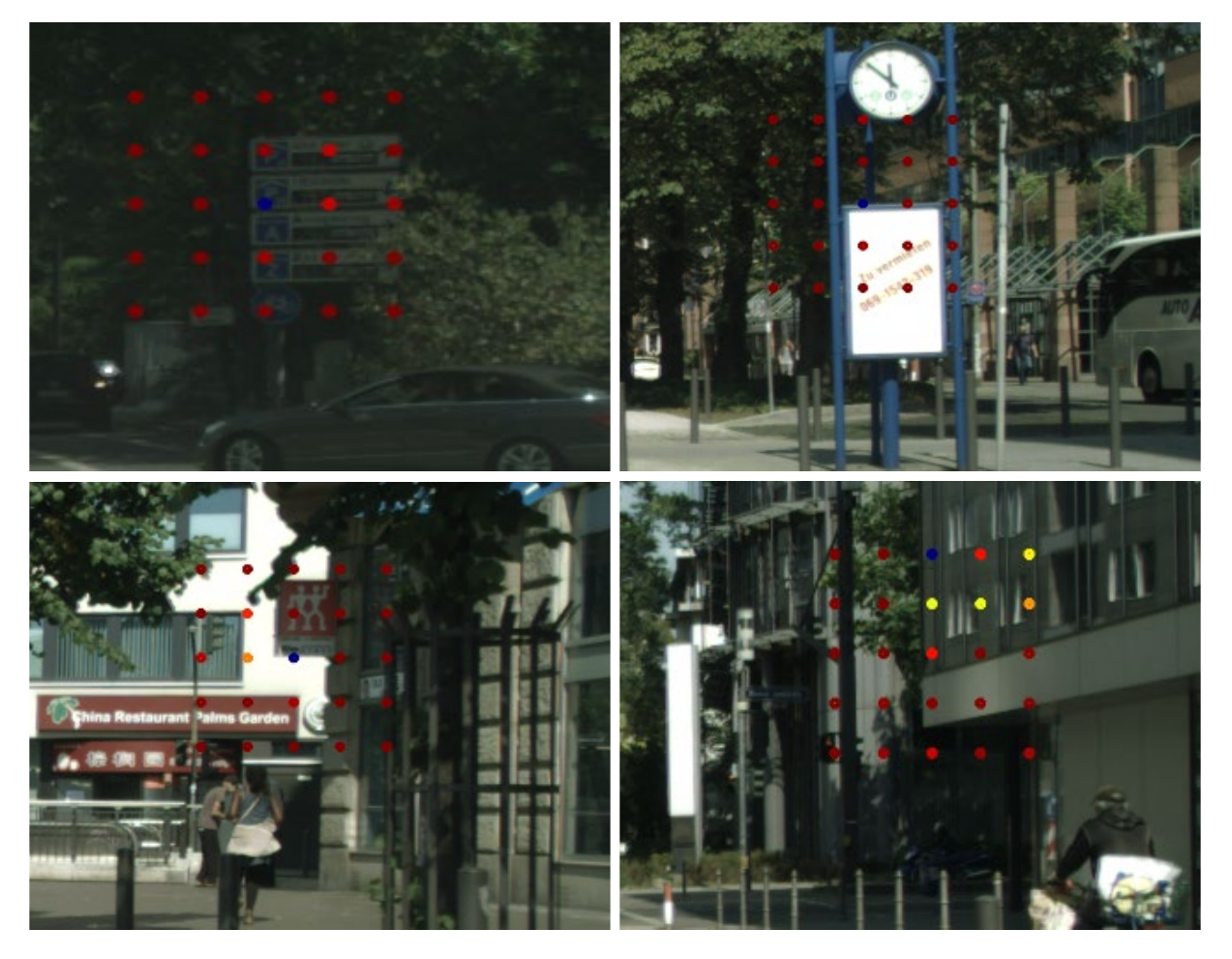}
	\caption{
		Visualization of local affinity map.
		(Best viewed in color)}
	\label{fig:example_results_on_local_sampled_point}
\end{figure}

\begin{figure}
	\centering
	\includegraphics[width=1.0\linewidth]{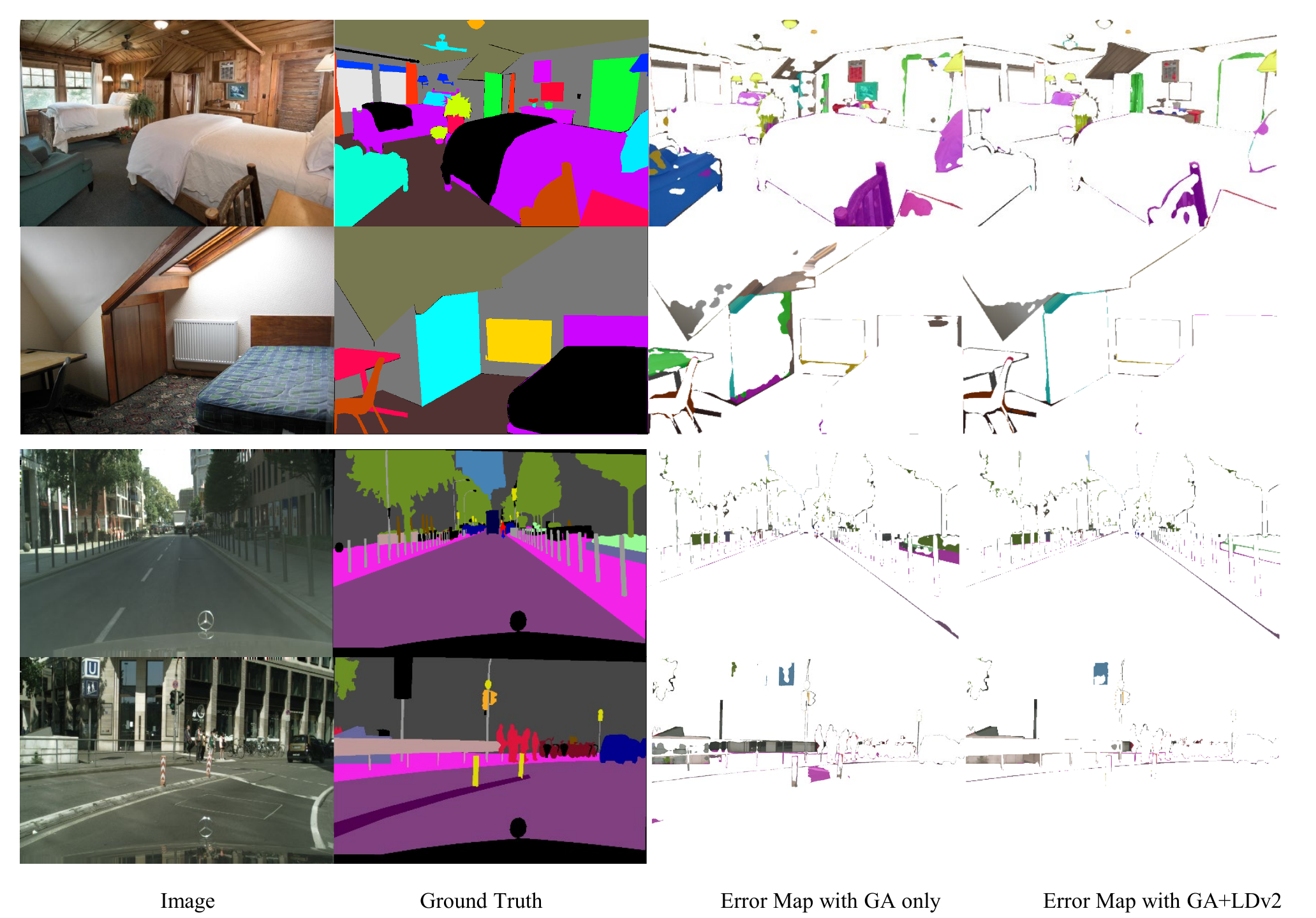}
	\caption{
		Visualization of error map on Cityscapes and ADE dataset over GA baseline.  
		(Best viewed in color)}
	\label{fig:example_results_on_error_map}
\end{figure}

\begin{figure*}
	\centering
	\includegraphics[width=0.80\linewidth]{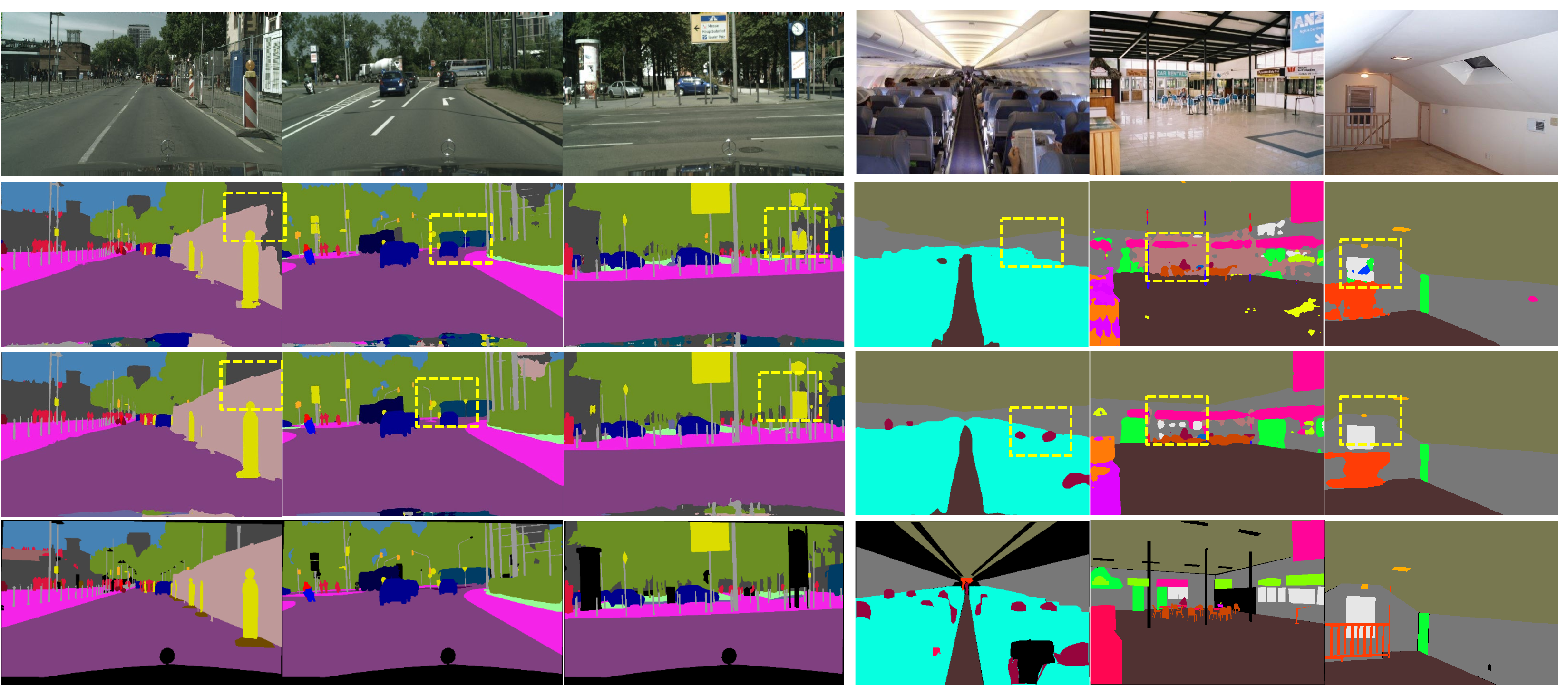}
	\caption{
		Example results of different methods. The images in each column from top to down are: (1) Input image. (2) Only GA heads (ASPP module is used). (3) With LDv2. (4) Ground truth. (Best viewed in color).}
	\label{fig:example_results_over_ga_head}
\end{figure*}

\subsubsection{Boundary improvement}
Since both our LD modules focus on obtaining the fine masks, we also show the effectiveness of our proposed LDv2 module on object boundaries. 
More specifically, we evaluate the performance of predicted mask boundaries, where we report the mean F-score of $19$ classes at 4-different thresholds following the recent GSCNN~\cite{gated-scnn}. This metric computes the F-score along the boundary of the predicted mask, given a small slack in distance. In our experiments, we use thresholds 0.00088, 0.001875, 0.00375, and 0.005 which correspond to 3, 5, 9, and 12 pixels respectively. For fair comparison, we only report models trained on the Cityscapes fine set. For all models, inference is done on a single-scale with full images as inputs.
From Table~\ref{tab:cityscapes_results_detail_f_score} we conclude that our methods improve the baseline object boundaries by a significant margin and are also slightly better than both GSCNN~\cite{gated-scnn} and Deeplabv3+~\cite{deeplabv3p} in different cases with four different thresholds. Similarly, our models perform considerably better, outperforming the two baselines including deeplabv3+~\cite{deeplabv3p} and G-SCNN~\cite{gated-scnn} by nearly 3\% to 4\% in the strictest regime.
This indicates the effectiveness of our modules on edge preservation.

\subsubsection{Efficiency Analysis} 
We carry out the efficiency analysis for both modules. In particular, we use the ASPP module as our GA head baseline and report two metrics including inference time and parameter number. Table \ref{tab:efficiency} shows the results. Our modules only bring a little inference time and parameter overhead compared with GA head baseline which proves the efficiency of our method. Meanwhile, in order to prove the fact that the gain is not brought by increased parameters and computation cost, we also include stronger backbone ResNet101 with GA head as reference in the last row. Our method achieves better performance improvement compared to GA modules with stronger backbone.

\subsection{Visualization Analysis}
\noindent \textbf{Visualization of GALD framework:} To give an intuitive understanding of our GALD framework, we add another two segmentation heads on features outputted from FCN and GA respectively.
The model is fine tuned until converge to analyze segmentation ability of features at different stages. 
Here we use LDv1 as examples.
Figure~\ref{fig:gald_mechanism} compares the segmentation results, segmentation based on GA resolves the ambiguities in FCN features but also tends to over smoothing regions of small patterns (see red boxes). Segmentation by GALD keeps the global structure of GA while refining back the details.

\noindent \textbf{Visualization of Local Affinity Map on Sampled Points:} To better show that the local context shares different global context adaptively, we visualize the local affinity map on one specific point. As shown in the first row of Figure~\ref{fig:example_results_on_local_sampled_point}, the points on the same object share the similar color as points on the tree in deep red color, and points on the traffic sign in light red color; This means the global context are distributed equally to each class. As shown in the second row, the points on the building along the windows are in different colors. That means each point shares the different global context since they have the appearance variation locally.

\noindent \textbf{Visualization of Error Map:} 
Figure~\ref{fig:example_results_on_error_map} gives error map on both Cityscapes~\cite{Cityscapes} and ADE20k~\cite{ADE20K} validation datasets using ASSP as GA head baselines. In particular, we use ResNet101 backbone as a strong baseline and LDv2 as the LD module. As shown in the last two columns, our method can better resolve the error around object boundaries and fine-grained information.

\noindent \textbf{Visualization with More Segmentation Results:}
Figure~\ref{fig:example_results_over_ga_head} gives more visual examples on Cityscape and ADE-20k validation datasets. For both datasets, our methods can obtain more fine-grained segmentation results. As shown in the yellow boxes, our method can obtain better object boundaries along the wall and brand
(see the first and second columns). Our method can also prohibit some false positives in the scene as shown in the second and last columns for uncertain things such as car on the truck, poles on the window. Finally, our method can find small missing objects (see the fourth row).

\subsection{Comparison to State-of-the-arts}

\begin{table*}[t!]
	\centering 
	\small
	\addtolength{\tabcolsep}{0pt}
	\caption{\small{
			Per-category results on the Cityscapes test set. Note that all the models are trained with \textbf{only fine annotated data}. Our methods outperform existing approaches and achieves \textbf{83.2}\% in mIoU. {All the models use ResNet101 as backbone.} }
	}
	\label{tab:cityscapes_results_detail_fine}
	\resizebox{\textwidth}{!}{
		\begin{tabular}{l|c|c|ccccccccccccccccccc}
			\toprule[0.2em]
			Method &  {Param(M)} & mIoU  & road & swalk & build. & wall & fence & pole & tlight & sign & veg & terrain & sky & person & rider & car & truck & bus & train & motor & bike \\
			\toprule[0.2em]
			PSPNet~\cite{pspnet}& 42.1 & 78.4 & 98.6 & 86.2 & 92.9 & 50.8 & 58.8 & 64.0 & 75.6 & 79.0 & 93.4 & 72.3 & 95.4 & 86.5 & 71.3 & 95.9 & 68.2 & 79.5 & 73.8 & 69.5 & 77.2 \\
			AAF~\cite{aaf} & - &79.1 & 98.5 & 85.6 & 93.0 & 53.8 & 58.9 & 65.9 & 75.0 & 78.4 & 93.7 &
			72.4 & 95.6 & 86.4 & 70.5 & 95.9 & 73.9 & 82.7 & 76.9 & 68.7 & 76.4\\
			DenseASPP~\cite{denseaspp} & 35.7 & 80.6 & {98.7} & 87.1 & 93.4 & 60.7 & 62.7 & 65.6 & 74.6 & 78.5 & 93.6 & 72.5 & 95.4 & 86.2 & 71.9 & 96.0 & 78.0 & 90.3 & 80.7 & 69.7 & 76.8 \\
			
			CCnet~\cite{ccnet} & 66.5 & 81.4 & - & -& -& -& -& -& -& -& -& -& -& -& -& -& -& -& -& - & -\\
			DANet~\cite{DAnet} & 66.6 & 81.5 & 98.6 & 87.1 & 93.5 & 56.1 & {63.3} & 69.7 & 77.3 & 81.3 & 93.9 & 72.9 & 95.7 & 87.3 & 72.9 & 96.2 & 76.8 & 89.4 & 86.5 & 72.2 & 78.2 \\
			
			SypGR ~\cite{SpyGR} & 63.5 &81.6 &98.7  & 86.9 &93.6 &57.6 &62.8& 70.3 &78.7 &81.7 &93.8& 72.4 &95.6& 88.1& 74.5 &96.2& 73.6 &88.8& 86.3 &72.1 &79.2 \\
			
			SPNet~\cite{strip_pooling} & - &82.0 & - & -& -& -& -& -& -& -& -& -& -& -& -& -& -& -& -& - & - \\	
			DGCNet~\cite{zhangli_dgcn} & 66.2 & 82.0 &98.7 &87.4 &93.9 &62.4 &\bf{63.4} &70.8& 78.7 &81.3& 94.0 &73.3 &95.8 &87.8& 73.7 &96.4 &76.0& 91.6& 81.6& 71.5& 78.2 \\
			DMNet~\cite{DGM_net} & - & 81.6 & - & -& -& -& -& -& -& -& -& -& -& -& -& -& -& -& -& - & - \\
			
			HANet~\cite{HAnet} & - & 82.1 &98.8 &88.0 &93.9& 60.5 &63.3 &71.3 &78.1& 81.3 &94.0 &72.9 &96.1 &87.9 &74.5 &96.5 &77.0& 88.0& 85.9 &72.7 &79.0 \\ 
			
			OCRNet~\cite{OCRnet} & - & 81.8  & - & -& -& -& -& -& -& -& -& -& -& -& -& -& -& -& -& - & - \\
			
			GFFNet~\cite{xiangtl_gff} & 70.5 &{82.3}  & {98.7} & {87.2} & {93.9} & 59.6 & {64.3} & \bf{71.5} & {78.3} & {82.2} & {94.0} & 72.6 & {95.9} & {88.2} & {73.9} & {96.5} & \bf{79.8} & {92.2} & 84.7 & 71.5 & {78.8}  \\
			
			ACNet~\cite{Fu_2019_ICCV}& - & 82.3 & 98.7 & 87.1 & 93.9&  61.6&  61.8&  71.4 & \bf{78.7} & 81.7&  94.0&  73.3&  96.0&  \bf{88.5} & 74.9 & 96.5 & 77.1 & 89.0&  89.2 & 71.4 & \bf{79.0} \\
			
			\hline
			Ours (ResNet101)-GALDv1 & {65.3} & {81.8} & {98.7} & {87.2} & {93.9} & {62.1} & 62.9 & {71.2} & {78.5} & {81.8} & {94.0} & {73.3} & \bf{96.0} & {88.1} & {74.4} & {96.5} & {79.4} & {92.5} & {89.8} & \bf{73.3} & {78.7} \\
			Ours (ResNet101)-GALDv2 & {66.8} & \bf{83.2} & \bf{98.8} & \bf{87.7} & \bf{94.0} & \bf{64.1} & 63.3 & {70.9} & {78.4} & \bf{82.1} & \bf{94.1} & \bf{73.7} & {95.9} & {88.3} & \bf{76.0} & \bf{96.5} & {79.0 }& \bf{93.8} & \bf{91.4} & {73.1} & {79.3} \\
			\hline
		\end{tabular}
	}
	\vspace{2mm}
\end{table*}

\begin{table*}[t!]
	\centering 
	\small
	\caption{ 
		\small Per-category results on the Cityscapes test set. Note that G-SCNN and our method are trained with \textbf{both fine annotated and coarse annotated data}. We achieve the state-of-the-art results with \textbf{83.5} mIoU. Best view on screen and zoom in.
	} 
	\addtolength{\tabcolsep}{0pt}
	\resizebox{\textwidth}{!}{
		\begin{tabular}{l|c|c|ccccccccccccccccccc}
			\toprule[0.2em]
			Method & Coarse & mIoU &  road & swalk & build. & wall & fence & pole & tlight & sign & veg & terrain & sky & person & rider & car & truck & bus & train & motor & bike \\
			\toprule[0.2em]
			PSP-Net \cite{pspnet} & \checkmark & 81.2 & 98.7 & 86.9 &  93.5 & 58.4 & 63.7 &  67.7 & 76.1 & 80.5 & 93.6 & 72.2  & 95.3  & 86.8 & 71.9 & 96.2 & 77.7 & 91.5 & 83.6 & 70.8 & 77.5 \\ 
			DeepLabV3 \cite{deeplabv3} & \checkmark & 81.3 & 98.6 & 86.2 & 93.5 & 55.2 & 63.2 & 70.0& 77.1 & 81.3 & 93.8    & 72.3&   95.9  &  87.6  & 73.4& 96.3 & 75.1 &  90.4  & 85.1 & 72.1 & 78.3 \\
			DeepLabV3+ \cite{deeplabv3p} & \checkmark & 81.9 & 98.7 & 87.0  & 93.9 & 59.5 & 63.7 & 71.4 &78.2 & 82.2 & 94.0& 73.0 & 95.8&88.0&  73.3 & 96.4  &  78.0 & 90.9 & 83.9 & 73.8 & 78.9 \\
			AutoDeepLab-L \cite{auto-deeplab} & \checkmark & 82.1 & {98.8} & \bf{87.6}  & 93.8  & 61.4  & 64.4  & 71.2 & 77.6 & 80.9 & \bf{94.1} & 72.7 & \bf{96.0} & 87.8 &  72.8  & 96.5  & 78.2 & 90.9 & 88.4 & 69.0 & 77.6 \\
			DPC \cite{DPC} & \checkmark & 82.7 & 98.7 & 87.1 & 93.8 & 57.7 & 63.5 & 71.0 & 78.0 & 82.1 &94.0 & 73.3 & 95.4 & 88.2 &  {74.5}  & {96.5}  &  {81.2} & {93.3} & {89.0} & \bf{74.1} & 79.0 \\
			\hline       
			Ours (ResNet101)-GALDv1 & \checkmark & 82.8  & \bf{98.8} & 87.5 & 94.0 & {65.3} & \bf{66.2} & 71.0 & 77.6 & 81.0 & 94.0 & 72.6 & 95.9 & 87.6 & 75.0 & 96.3 & 80.0 & 90.3 & 87.9 & 72.9 & 78.9 \\
			Ours (ResNet101)-GALDV2 & \checkmark & \bf{83.5} & 98.7 & 87.3 & \bf{94.2} & \bf{65.7} & 65.6 & \bf{72.3} & \bf{78.5} & \bf{82.3} & 94.0 & \bf{74.1} & 95.9 & \bf{88.4} & \bf{76.3} & \bf{96.6} & \bf{81.2} & \bf{93.3} & \bf{90.1} & 73.5 &  \bf{79.4}  
			\\
			\hline
		\end{tabular}
	}
	\vspace{2mm}
	
	\label{tab:cityscapes_results_coarse_data}
\end{table*}

\begin{table}[!t]
	\centering
	\footnotesize
	\caption{\footnotesize State-of-the-art comparison on Pascal Context test set. }
	\setlength{\tabcolsep}{2.0 pt}
	\begin{tabular}{l|c|c}
		\toprule[0.2em]
		Method & Backbone & mIoU (\%)   \\
		\toprule[0.2em]
		RefineNet~\cite{refinenet} & ResNet152 &  47.3 \\
		PSPNet~\cite{pspnet} & ResNet101 & 47.8 \\
		EncNet~\cite{context_encoding} & ResNet101   &51.7  \\
		Ding~\textit{et al.}~\cite{ding2018context} & ResNet101  &51.6  \\
		DANet~\cite{DAnet}& ResNet101  & 52.6   \\
		SGR~\cite{SGR_gcn} & ResNet101 & 52.5 \\
		ANN~\cite{annet} & ResNet101 & 52.8 \\
		SpyGR~\cite{SpyGR} & ResNet101 & 52.8 \\ 
		BFPNet~\cite{BAFPNet} & ResNet101 & 53.6 \\
		EMANet~\cite{2019Expectation} & ResNet101 & 53.1 \\
		GFFNet~\cite{xiangtl_gff} & ResNet101 & {54.2} \\
		SPnet~\cite{strip_pooling} & ResNet101 &  54.5 \\
		APCnet~\cite{APCNet} & ResNet101 &  54.7 \\
		\hline
		{Ours (GALDv1)} & ResNet-101 &\bf 53.5 \\
		{Ours (GALDv2)} & ResNet-101 &\bf 55.2 \\
		\hline 
	\end{tabular}
	\label{tab:pascal_context}
\end{table}

\begin{table}[!ht]
	\centering
	\footnotesize
	\setlength{\tabcolsep}{0.6 pt}
	\caption{\footnotesize{State-of-the-art comparison experiments on ADE20K validation set. Our models achieve top performance measured by both mIoU and pixel accuracy.} }
	\begin{tabular}{l|c|c|c}
		\toprule[0.2em]
		Method & Backbone & mIoU($\%$) & Pixel Acc.(\%) \\
		\toprule[0.2em]
		PSPNet~\cite{pspnet} & ResNet101 &  43.29 & 81.39\\ 
		PSANet~\cite{psanet} & ResNet101 &  43.77 & 81.51\\ 
		EncNet~\cite{context_encoding}  & ResNet101 & 44.65 & 81.69
		\\ 
		GCUNet~\cite{beyond_grids} & ResNet101 & 44.81 & 81.19\\
		OCRNet~\cite{OCRnet} & ResNet101 & 45.28 & - \\
		ANNet~\cite{annet} & ResNet101 & 45.24 & - \\ 
		CCNet~\cite{ccnet} & ResNet101 & 45.22 & - \\
		CFNet~\cite{co_current_net}  & ResNet101 & 44.89 & - \\
		SPnet~\cite{strip_pooling} &ResNet101  & 45.60 & 82.01 \\
		ACnet~\cite{Fu_2019_ICCV} & ResNet101 & 45.90 & 81.96 \\
		\hline
		CGNL only & ResNet101 & 43.65 & 80.13 \\
		Ours (GALDv1) & ResNet101 & \textbf{44.95} & 81.33 \\
		ASPP only  & ResNet101 & {43.85} & 81.21 \\
		Ours (GALDv2) & ResNet101 & \textbf{46.38} & \textbf{82.22} \\ 
		\hline
	\end{tabular}
	
	\label{tab:ade20k_res}
\end{table}

\begin{table}[!ht]
	\centering
	\footnotesize
	\setlength{\tabcolsep}{0.6pt}
	\caption{\footnotesize{State-of-the-art comparison experiments on Camvid test set. Our models achieve top performance in both different settings.} }
	\begin{tabular}{ l|c|c|c}
		\toprule[0.2em]
		Method & Backbone & Pretrain & mIoU(\%)\\
		\toprule[0.2em]
		DenseDecoder~\cite{densedecoder} & ResNext-101 & ImageNet & 70.9 \\
		BAFPnet~\cite{BAFPNet} & ResNet-101 & ImageNet & 74.1 \\
		Ours (GALDV1) & ResNet-101 & ImageNet & \textbf{74.9}  \\
		Ours (GALDV2) & ResNet-101 & ImageNet & \textbf{76.5}  \\
		\hline
		VideoGCRF~\cite{VideoGCRF} & ResNet-101 & Cityscapes & 75.2 \\
		Video Propagation~\cite{video_propagation} & Wider-ResNet & Cityscapes & 79.8 \\
		Ours (GALDV1) & ResNet-101 & Cityscapes & \textbf{80.3} \\
		Ours (GALDV2) & ResNet-101 & Cityscapes & \textbf{81.6} \\
		\hline
	\end{tabular}
	
	\label{tab:camvid_res}
\end{table}

\begin{table}[!ht]
	\centering
	\footnotesize
	\caption{\footnotesize{State-of-the-art comparison experiments on COCO-stuff validation set. Our models achieve top performance measured by mIoU.} }
	\begin{tabular}{l|l|l}
		\toprule[0.2em]
		Method & BackBone & mIoU(\%) \\
		\toprule[0.2em]
		RefineNet~\cite{refinenet} & ResNet101 & 33.6\\
		DSSPN~\cite{DSSPN} & ResNet101 & 36.2 \\
		CCLNet~\cite{ding2018context} & ResNet101 & 35.7 \\
		DANet~\cite{DAnet} & ResNet101 &  37.2 \\
		GFFNet~\cite{xiangtl_gff} & ResNet101 & 39.2 \\
		SPyGR~\cite{SpyGR} & ResNet101 & 39.9 \\
		OCR~\cite{OCRnet} & ResNet101 & 39.5 \\
		\hline
		Ours (GALDv1) & ResNet101 & 38.5 \\
		Ours (GALDv2) & ResNet101 & \bf{40.1} \\
		\hline
	\end{tabular}
	
	\label{tab:coco_stuff_res}
\end{table}

\noindent \textbf{Results on Cityscapes dataset:}
We first compare our models with other state-of-the-art methods. We choose dilated  ResNet101 as backbone for fair comparison. In particular, we consider two settings: using fine annotation data only and using fine annotation data and coarse data jointly. The models are tested with multi-scale inference with flip operation. For fine dataset, we train our models for 300 epochs with both train and validation data at initial learning rate 0.01.
For the first setting, we report our results in Table~\ref{tab:cityscapes_results_detail_fine}. Our GALDv2 surpasses all previous methods. In particular, GALDv2 achieves 83.2\% mIoU with a large margin compared to previous methods. {We also report several model size for reference in the second column.} To the best of our knowledge, this is the first single ResNet101 based model that surpasses 83\% mIoU on Cityscapes test server using \textbf{\em only fine annotated} data. Compared with ACnet~\cite{Fu_2019_ICCV}, the margin is 1\% mIoU. We report the second setting's results in Table~\ref{tab:cityscapes_results_coarse_data}.
Different from training on fine data set only, we set batch size 16 and fix batch normalization layers in our model for about 40 epochs with initial learning rate 0.001. Then we fine tune the model back on the fine data set for 20 epochs with learning rate 0.001. By using extra coarse annotation data in training, our method achieves 82.9\% mIoU and 83.5 \% mIoU, surpassing the previous state-of-the-art methods~\cite{DPC,auto-deeplab}.

\noindent \textbf{Results on ADE20k dataset:} 
We conduct experiments on ADE20k dataset to validate the effectiveness of our method on more challenging scenes. Following the previous works, we adopt data
augmentation with random scaling and multi-scale testing in training and testing. Specifically, we set the initial
learning rate to 0.01, crop size to 512 $\times$ 512, batch size to
16 and train models for 240 epochs. The model is trained
and tested with multi-scale schemes (0.5, 0.75, 1.0, 1.25,
1.5, 1.75). We evaluate our GALD models by the standard metrics of
pixel accuracy (pixAcc) and mean intersection of union (Mean IoU). Quantitative results are shown in Table~\ref{tab:ade20k_res}. Different from previous works focusing on global context modeling alone, our method models local context with global prior as guidance with extra advantages to handle small various objects in the scene. As a result, compared with GA baselines, our proposed LD modules obtain significant improvement by 1.3 \% and 2.7 \% mIoU respectively.

\noindent \textbf{Results on Pascal Context dataset:} We carry out experiments on the PASCAL Context dataset.
Specifically, we set the initial learning rate to 0.001, crop size to 532$\times$532, batch size to 16 and train models for 180 epochs. Quantitative results of PASCAL Context are shown in Table~\ref{tab:pascal_context}. GALDv2 with ResNet-101 achieves a Mean IoU of 55.3\%, which outperforms previous methods by a large margin. 

\noindent \textbf{Results on Camvid dataset:} We carry out experiments on Camvid dataset. Since this dataset is small, previous works use Cityscapes for pretraining to obtain stronger results.
Specifically, starting from ImageNet pretrained weights, we set the initial learning rate to 0.01, crop size to 640$\times$ 640, batch size to 16 and train the models for 180 epochs on Cityscapes.
Subsequently we lower down the initial learning rate to 0.001 and epochs to 90.
For a fair comparison, we compare both ImageNet pretrained and Cityscapes pretrained models. As shown in Table \ref{tab:camvid_res}, our methods achieve significant gains over other state-of-the-arts in both cases.

\noindent \textbf{Results on COCO-stuff dataset:} We finally conduct experiments on the COCO Stuff dataset to verify the generalization of our method. Specifically, we set the initial learning rate to 0.001, crop size to 532$\times$532, batch size to 16 and train models for 240 epochs. Compared with previous state-of-the-art in Table~\ref{tab:coco_stuff_res}, our method outperforms the existing methods by a clear margin. 

%% file: 5conclusion.tex
\section{Conclusion}

In this paper, we propose the GALD framework to adaptively distribute global information to each position for scene parsing task. In contrast to existing methods that assign global information uniformly to each position and cause the problem of blurring and missing details, we propose two different methods to distribute global information adaptively according pattern distributions over the image. One is by learning a set of mask maps to sharpen the large objects and the other is to estimate global-to-local affinity map adaptively to assign global context to each pixel location. Both modules are lightweight and efficient during the inference. In general, GALD benefits from both the GA module for ambiguity resolving and LD module for detail refinement. The detailed ablation and visualization experiments show that our method captures more fine-grained contextual information effectively and gives more precise segmentation results including object boundaries and finding missing small objects compared with strong baselines. Our proposed GALDnet achieves outstanding performance consistently on five scene segmentation datasets including Cityscapes, ADE-20k, Pascal Context, Camvid and COCO-stuff.